\documentclass[pdftex,10pt,a4paper]{article}
\usepackage{afterpage}
\usepackage{amsmath}
\usepackage{amsfonts}
\usepackage{amssymb}
\usepackage{txfonts} 
\DeclareMathAlphabet{\mathpzc}{OT1}{pzc}{m}{it} 
\usepackage{bbm,bbold}
\usepackage{mathptmx}       
\usepackage{makeidx}         
\usepackage{graphicx}        
\graphicspath{{./}}
\usepackage{multicol}        
\usepackage{acronym}
\usepackage{multirow}
\usepackage{color, soul}

\usepackage[caption=false, font=footnotesize]{subfig}
\usepackage{setspace}

\usepackage{footnote}
\usepackage{tablefootnote}
\makesavenoteenv{tabular}
\makesavenoteenv{table}

\usepackage{nameref}
\usepackage{natbib}
\DeclareMathAlphabet{\mathcal}{OMS}{cmsy}{m}{n}
\SetMathAlphabet{\mathcal}{bold}{OMS}{cmsy}{b}{n}
\usepackage{mwe}

\title{An iterative scheme for feature-based positioning using a weighted dissimilarity measure}

\author{Caifa Zhou (caifa.zhou@geod.baug.ethz.ch), Andreas Wieser}
\begin{document}
	\maketitle
\acrodef{vc}[VC]{Vapnik-Chervonenkis}
\acrodef{knn}[$ k $NN]{$ k $ nearest neighbors}
\acrodef{lbs}[LBS]{location-based service}
\acrodef{ilbs}[ILBS]{indoor location-based service}
\acrodef{fips}[FIPS]{fingerprinting-based indoor positioning system}
\acrodef{fbp}[FbP]{feature-based positioning}
\acrodef{gnss}[GNSS]{global navigation satellite system}
\acrodef{ips}[IPS]{indoor positioning system}
\acrodef{rfid}[RFID]{radio frequency identification}
\acrodef{uwb}[UWB]{ultra wideband}
\acrodef{wlan}[WLAN]{wireless local area network}
\acrodef{rss}[RSS]{received signal strength}
\acrodef{ap}[AP]{access point}
\acrodef{roi}[RoI]{region of interest}
\acrodef{lasso}[LASSO]{least absolute shrinkage and selection operator}
\acrodef{rfm}[RFM]{reference fingerprint map}
\acrodef{rfm1}[RFM]{reference feature map}
\acrodef{map}[MAP]{maximum a posteriori}
\acrodef{cpa}[CPA]{cumulative positioning accuracy}
\acrodef{mse}[MSE]{mean squared error}
\acrodef{tp}[TP]{test position}
\acrodef{wrt}[w.r.t.]{with respect to}
\acrodef{mac}[MAC]{media access control}
\acrodef{imu}[IMU]{inertial measurement unit}
\acrodef{foba}[adaFoBa]{adaptive forward-backward greedy}
\acrodef{rp}[RP]{reference point}
\acrodef{mji}[MJI]{modified Jaccard index}
\acrodef{ble}[BLE]{Bluetooth low energy}
\acrodef{oil}[OIL]{organic indoor localization}
\acrodef{will}[WILL]{wireless indoor localization without site survey}
\acrodef{hiwl}[HIWL]{hidden Morkov model-based indoor wireless localization}
\acrodef{svm}[SVM]{support vector machine}
\acrodef{lda}[LDA]{linear discriminant analysis}
\acrodef{sop}[SoP]{signal of opportunity}
\acrodef{ks}[KS]{Kolmogorov-Smirnov}
\acrodef{ecdf}[ECDF]{empirical cumulative distribution function}
\acrodef{cdm}[CDM]{compound dissimilarity measure}
\acrodef{rmse}[RMSE]{root mean squared error}
\acrodef{rcdm}[RCDM]{relatively weighted compound dissimilarity measure}
\acrodef{acdm}[ACDM]{averagely weighted compound dissimilarity measure}
\acrodef{cv}[CV]{cross validation}
\acrodef{pdf}[PDF]{probability density function}
\acrodef{laafu}[LAAFU]{localization with altered APs and fingerprint update}
\acrodef{amd}[AMD]{average mutual distance}
\acrodef{rpc}[RPC]{recall precision curve}
\acrodef{auc}[AUC]{area under the ROC curve}
\acrodef{bo}[BO]{Bayesian optimization}
\acrodef{roc}[ROC]{receiver operating characteristic}
\acrodef{rocNick}[ROC]{relative operating characteristic curve}
\acrodef{svr}[SVR]{support vector regression}
\acrodef{smbo}[SMBO]{sequential model-based optimization}
\acrodef{iou}[IoU]{intersection over union}
\acrodef{gp}[GP]{Gaussian process}
\acrodef{ei}[EI]{expected improvement}
\acrodef{tpr}[TPR]{true positive rate}
\acrodef{fpr}[FPR]{false positive rate}
\acrodef{ks}[KS]{kernel smoothing}
\acrodef{rkhs}[RKHS]{reproducing kernel Hilbert space}
\acrodef{qcd}[QCD]{querying-based change detection}
\acrodef{mae}[MAE]{mean absolute error}
\acrodef{kde}[KDE]{kernel-based density estimation}
\acrodef{pdr}[PDR]{pedestrian dead reckoning}
\acrodef{rsm}[RSM]{random subspace method}
\acrodef{ransac}[RANSAC]{random sampling consensus}
\acrodef{dbscan}[DBSCAN]{density based spatial clustering of applications with noise}
\acrodef{amise}[AMISE]{asymptotic mean integrated squared error}
\acrodef{epdf}[EPDF]{empirical probability density function}
\acrodef{cr}[CR]{confidence region}
\acrodef{ce}[CE]{circular error}
\acrodef{tp}[TP]{true positive}
\acrodef{fn}[FN]{false negative}
\acrodef{tn}[TN]{true negative}
\acrodef{fp}[FP]{false positive}
\acrodef{tpr}[TPR]{true positive rate}
\acrodef{tnr}[TNR]{true negative rate}
\acrodef{plsr}[PLSR]{partial least square regression}
\acrodef{gpr}[GPR]{Gaussian process regression}
\acrodef{std}[STD]{standard deviation}
\acrodef{bic}[BIC]{Bayesian information criterion}
\acrodef{ugv}[UGV]{unmanned ground vehicle}
\acrodef{ldplm}[LDPLM]{Log-distance path loss model}
\acrodef{slam}[SLAM]{simultaneous localization and mapping}
\acrodef{ekf}[EKF]{extended Kalman filter}
\acrodef{mad}[MAD]{median absolute deviation}
\acrodef{mle}[MLE]{maximum likelihood estimation}
\acrodef{mcd}[MCD]{minimum covariance determinant}
\acrodef{tf}[TF]{termination flag}
\newcommand{\todo}[1]{\btext{\textit{To be done: #1}}}
\newcommand{\tocite}{\btext{[CITE] }}
\newcommand{\toref}{\btext{[REF-FIG-TAB] }}
\newcommand{\appRef}{Appendix }
\newcommand{\secPref}{Section }
\newcommand{\figPref}{Fig.}
\newcommand{\tabPref}{TABLE }
\newcommand{\eg}{e.g. }
\newcommand{\ie}{i.e. }
\newcommand{\etal}{et~al.}
\newcommand{\rhl}[1]{\textcolor{red}{\hl{#1}}}
\newcommand{\btext}[1]{\textcolor{blue}{{#1}}}
\newcommand{\colRev}[1]{\btext{#1}}
\newcommand{\colRevOff}[1]{{#1}}
\newcommand{\convSetSym}[1]{\varmathbb{#1}}
\newcommand{\setSym}[1]{\mathbbm{#1}}
\newcommand{\vecSym}[1]{\mathbf{#1}}
\newcommand{\funSym}[1]{\mathpzc{#1}}
\newcommand{\textSym}[1]{\mathrm{#1}}
\newcommand{\mapSym}{\mapsto}
\newcommand{\texSym}[1]{\mathrm{#1}}
\newcommand{\intSet}[1]{\{1, 2, \cdots, #1\}}
\newcommand{\setSymScript}[3]{\setSym{#1}_{#2}^{\texSym{#3}}}
\newcommand{\vecSymScript}[3]{\vecSym{#1}_{#2}^{\texSym{#3}}}
\newcommand{\norSymScript}[3]{#1_{#2}^{\texSym{#3}}}
\newcommand{\compComp}[1]{\mathcal{O}(#1)}
\newcommand\blankpage{%
	\null
	\thispagestyle{empty}%
	\addtocounter{page}{-1}%
	\newpage}
\newlength{\tempdima}
\newcommand{\rowname}[1]
{\rotatebox{90}{\makebox[\tempdima][c]{{#1}}}}

\renewcommand{\thesubfigure}{\alph{subfigure}}
\newcommand{\mycaption}[1]
{\refstepcounter{subfigure}\textbf{(\thesubfigure) }{\ignorespaces #1}}

\begin{abstract}
	We propose an iterative scheme for \acl{fbp} using a new weighted dissimilarity measure with the goal of reducing the impact of large errors among the measured or modeled features. The weights are computed from the location-dependent standard deviations of the features and stored as part of the \acl{rfm} (\acs{rfm}). Spatial filtering and kernel smoothing of the kinematically collected raw data allow efficiently estimating the standard deviations during \acs{rfm} generation. In the positioning stage, the weights control the contribution of each feature to the dissimilarity measure, which in turn quantifies the difference between the set of online measured features and the fingerprints stored in the \acs{rfm}. Features with little variability contribute more to the estimated position than features with high variability. Iterations are necessary because the variability depends on the location, and the location is initially unknown when estimating the position. Using real WiFi signal strength data from extended test measurements with ground truth in an office building, we show that the standard deviations of these features vary considerably within the region of interest and are neither simple functions of the signal strength nor of the distances from the corresponding access points. This is the motivation to include the empirical standard deviations in the \acs{rfm}. We then analyze the deviations of the estimated positions with and without the location-dependent weighting. In the present example the maximum radial positioning error from ground truth are reduced by 40\% comparing to \acs{knn} without the weighted dissimilarity measure.
	
\end{abstract}

\section{Introduction}\label{sec:intro}
Feature-based (\ie fingerprinting-based) indoor positioning systems (\acsp{fips}), one of the promising indoor positioning solutions, have been proposed using various types of features (\eg \acs{wlan}/\acs{ble} signal strengths \citep{Padmanabhan2000,Youssef2008,zhuang2016smartphone}, geomagnetic field strengths \citep{he2018geomagnetism} or visible patterns \citep{guan2016vision}) for providing indoor \acfp{lbs} to pedestrians \citep{brena2017evolution,he2016wi,pei2016survey}. The positioning accuracy of the state-of-the-art \acsp{fips} using the \acf{rss} of \acs{wlan} \acfp{ap} is in the range of a few meters \citep{mautz2012indoor}. This is adequate for pedestrian indoor positioning and navigation in many cases. However, unexpected and unacceptably large errors (\eg $>$ 20~m in horizontal coordinates \citep{torres2017smartphone}) can be observed in real environments. They jeopardize the practical usability of \acsp{fips} \citep{wu2017mitigating,torres2017analysis}. Such large errors may be caused by large deviations of the measured or stored feature values when performing the location estimation \citep{KAEMARUNGSI2012292}.

In order to benefit from the attractive characteristics of \acs{fips} while mitigating large errors, the trend is to combine the feature-based positioning with other techniques. Such hybrid approaches combine the feature-based information with \eg \acf{pdr} \citep{li2016integrated}, map matching \citep{wang2015floor,wang2012no} or infrared ranging \citep{bitew2015hybrid}. In addition, Bayes filtering methods, such as Kalman filters or particle filters are used to improve the estimated trajectory of pedestrians by combining the measurements with assumptions on the user's motion \citep{li2016integrated,robesaat2017improved}. Merging different positioning solutions may help mitigating the impact of large errors of individual observations on the quality of a specific type of \acsp{lbs}. However, such approaches requires either deploying additional infrastructure or providing extra information (\eg the indoor map). It would be useful to detect or mitigate large errors in \acs{fips} using only intrinsically available data. This has attracted little research attention in the past, see \eg\citep{wu2017mitigating,torres2017analysis,lemic2019regression}, and is the motivation for the present contribution.

We base our approach on the variability of the feature values at each individual location. Feature values measured during the positioning stage are snapshots affected by noise. Even if the expected value of the feature has not changed since the data collection for the generation of the \acf{rfm}, the measured value may be closer to the \acs{rfm} value at a different position than to the one at the correct position because of this noise. It is therefore important to take the noise into account when assessing the similarity of measured and stored feature values. We facilitate this by storing the empirical \acfp{std} in the \acs{rfm} which is generated during the offline phase for representing the relationship between locations and their associated features. The estimation of the variability is carried out by empirically analyzing the spatial distribution of the raw data (\eg \acs{rss} values) included in the \acs{rfm}. It yields an extended representation of the \acs{rfm}, which contains not only the spatially smoothed feature values, but also the location-wise estimated \acs{std} of each individual feature (see \secPref\ref{sec:robust_var}). These values can then be used to mitigate the impact of large errors in \acs{fips}. To this end we propose a weighted dissimilarity measure, which quantifies the difference between the online measured features and the features stored in the \acs{rfm}, by adapting the contribution of the individual features to the dissimilarity measure relative to their estimated \acs{std} values (see \secPref\ref{subsec:wdm}). The positioning process is carried out in an iterative way because we need to assume the user's location, which is required for retrieving the \acs{std} of the online measured features (see \secPref\ref{subsec:iter_sch}). Beyond the use further discussed in this paper, the location-dependent standard deviations can also be employed for identifying (large) changes of features which may need an update of the \acs{rfm}, see e.g.~\citep{he2016wi,8445663,7676155}.

The remaining of the paper is organized as follows: \secPref\ref{sec:related} summarizes the work related to reducing large errors in an \acs{fips}. The fundamentals of the \acl{fbp} are briefly described in \secPref\ref{sec:fbp}. The robust estimation of the variability of the \acs{rfm} and its application to positioning are presented in \secPref\ref{sec:robust_var} and \ref{sec:iter_pos}, respectively. Finally, the evaluation of the variability estimation as well as the positioning performance using the iterative scheme are presented in \secPref\ref{sec:perf_ana} for a real world dataset.

%
\section{Related work}\label{sec:related}
Herein we focus on publications that address the detection and reduction of large errors in an \acs{fips}. We refer the interested readers to \citep{mautz2012indoor,he2016wi,brena2017evolution} for more general information about indoor positioning. A comprehensive comparison of different feature-based indoor positioning algorithms using various similarity/dissimilarity metrics is available in \citep{retscher2016,TORRESSOSPEDRA20159263,8115922}. A short review of the methods used for generation or creation of the \acs{rfm} can be found in \eg \cite{he2016wi,Zhou2018}.

\cite{torres2017analysis} provides a detailed analysis of the sources of large errors when employing deterministic \acl{fbp} approaches (\eg \acs{knn}). The analysis is based on simulations for different indoor scenarios. The authors consider the influence of several factors such as the quantization error of signal acquisition, the density of the reference measurements, and the selected dissimilarity metrics on the positioning error. The analysis shows that large observation errors mostly occur at locations where both the mean and the maximum value of the \acs{rss} are low. However, the authors do not report about a validation of their analysis in a really deployed \acsp{fips}. On a related note, \cite{KAEMARUNGSI2012292} proposes to simply disregard features with a large standard deviation for the estimation of the user's position. 

There are only few works that focus on reducing or estimating the positioning errors based on the analysis of the \acs{rfm}\footnote{\cite{torres2017analysis} provides a complete discussion of the works focusing on reducing large positioning errors by support of other technologies (\eg \acs{pdr},  or Bayes filtering).}. \cite{wu2017mitigating} introduces a weighted dissimilarity measure by computing the discriminative indicator for each feature according to the Log-distance path loss model. However, the variability of the online measured features which has an impact on the estimation of the discriminative factor is not taken into account. In \citep{lemic2019regression} and \citep{li2019wireless}, the authors propose different regression models (\eg neural networks, random forest, or Gaussian processes) for estimating the positioning errors and uncertainties that can be used to improve the performance of tracking a pedestrian's trajectory. Even if this is not the focus of these papers, the results suggest that the regression-based error prediction models cannot help to mitigate {\it large} errors because the predicted errors have a large uncertainty.

Compared to previous publications, we carry out the variability analysis of the \acs{rfm} using a kinematically collected dataset, which includes not only the noise originating from the short term fluctuations of the features measured by a mobile device, but also the noise introduced by the motion status (\eg moving speed and headings) of the mobile device. This setup is closer to the realistic situation of positioning and tracking pedestrians. The estimation of the variability is based solely on the raw \acs{rfm} and is later used for reducing large errors by introducing an iterative scheme with the weighted dissimilarity measure in the online positioning phase.

\section{Feature-based positioning}\label{sec:fbp}
We start this section by introducing the fundamental concepts of feature-based positioning and then briefly describe the process of kinematically collecting the \acs{rfm}.

\subsection{Fundamental concepts}\label{subsec:fbpProb}
Each measured feature is uniquely identifiable and has a measured value. For example, the signal from an \acs{ap}, can be identified by its \acl{mac} address and is associated with an \acs{rss}. Features are thus formulated as pairs of attribute $ a $ and value $ v $, \ie $ (a, v) $. A measurement (\ie fingerprint) $ \setSymScript{O}{i}{u} $ taken by the user $ \mathrm{u} $ at the location/time $ i $ consists of a set of measured features, \ie $ \setSymScript{O}{i}{u} := \{(\norSymScript{a}{ik}{u}, \norSymScript{v}{ik}{u}) | \norSymScript{a}{ik}{u}\in\setSym{A}; \norSymScript{v}{ik}{u}\in\convSetSym{R};k\in\intSet{\norSymScript{N}{i}{u}} \} $,  where $ \setSym{A} $ is the complete set of the identifiers of all available features and $ N_i^{\texSym{u}} $ ($ N_i^{\texSym{u}} = |\setSymScript{O}{i}{u}|$) is the number of features observed by the user $ \texSym{u} $ at $ i $. The set of attributes of $ \setSymScript{O}{i}{u} $ is defined as $ \setSymScript{A}{i}{u} := \{a_{ik}^{\texSym{u}}|\exists (a_{ik} ^{\texSym{u}}, v_{ik})\in \setSymScript{O}{i}{u}\} $ ($ \setSymScript{A}{i}{u} \subseteq \setSym{A}$). The positioning process consists of inferring the estimated user location $ \vecSymScript{\hat{l}}{i}{u}=\funSym{f}(\setSymScript{O}{i}{u}) (\vecSymScript{\hat{l}}{i}{u}\in\convSetSym{R}^{d}) $ as a function of the measurement and the \acs{rfm} $ \funSym{F} $, where $\funSym{f}$ is a suitable mapping algorithm from the measurement to location\footnote{ The \acs{rfm} $ \funSym{F} $ is omitted from the positioning algorithm $ \funSym{f} $ for simplicity. $ d $ (\eg $ d=2 $) is the dimension of the coordinates.}. $ \funSym{F} $ represents the relationship between the location $ \vecSym{l} $ and the measurement $ \setSym{O} $, \ie $ \funSym{F}:\vecSym{l}\mapSym\setSym{O} | \vecSym{l}\in\setSym{G} $ throughout the \acf{roi} $ \setSym{G} $. If the \acs{rfm} is discretely represented, we denote it as $ \setSym{F} := \{(\vecSym{l}_j, \setSymScript{\tilde{O}}{j}{})| \vecSym{l}_j\in \setSym{G}, j\in\intSet{|\setSym{F}|} \} $ (where $ \setSymScript{\tilde{O}}{j}{} = \funSym{F}(\vecSym{l}_j) $). A discrete \acs{rfm} can be obtained \eg by collecting fingerprints at different known or independently measured locations within the \acs{roi} $ \setSym{G}$.

\subsection{Kinematically acquired \acs{rfm}}
The kinematically obtained dataset used as the basis for the \acs{rfm} herein has already been employed in \citep{Zhou2018}. It was acquired using a mobile device (Nexus 6P) whose ground truth location was continuously measured with mm- to cm-level accuracy by a total station tracking a mini prism mounted on top of the mobile device. This procedure enables to simultaneously obtain accurate reference coordinates and the fingerprinting data collected by a pedestrian. The measurements were obtained at arbitrary locations lying on the trajectory of a pedestrian because the data acquisition on the mobile phone is passively triggered by the status of measurable features (\eg the arrival of new features or the change of feature values) \citep{Schulz:2018:SWT:3210240.3210333}. By carrying out a thorough site-survey, all the collected measurements and their tracked trajectories were merged and used to generate the raw \acs{rfm}. Herein we use this dataset as the basis of our analysis. More details of its acquisition and processing can be found in \citep{Zhou2018}.

\begin{figure}[!h]
	\centering
	\settoheight{\tempdima}{\includegraphics[width=.45\linewidth]{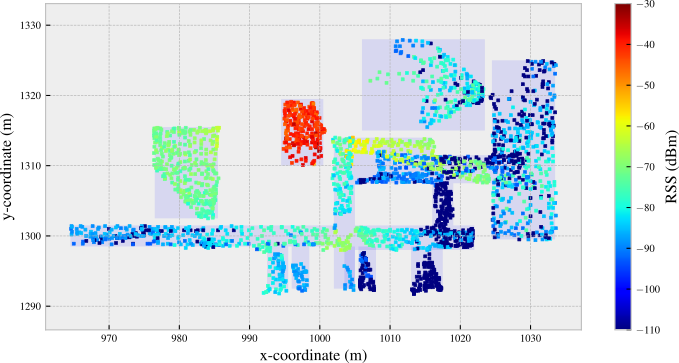}}
	\begin{tabular}{c@{ }c@{ }c@{ }}
		&\acs{ap}: \textit{9c:50:ee:09:5f:30} & \acs{ap}: \textit{9c:50:ee:09:61:d1}\vspace{-1.5ex}\\
		\rowname{Raw \acs{rfm}}&\hspace{2ex}\subfloat[]{\includegraphics[width=0.45\columnwidth]{rss_raw_mac_9c_50_ee_09_5f_30.png}
			\label{subfig:raw_rss_1}}&\hspace{2ex}
		\subfloat[]{\includegraphics[width=0.45\columnwidth]{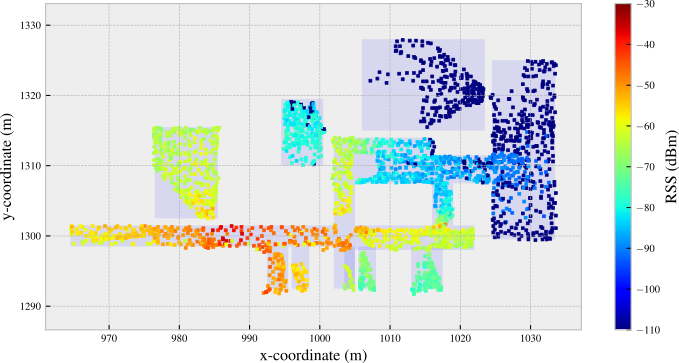}
			\label{subfig:raw_rss_2}}\vspace{-1ex}\\
		\rowname{Spatially filtered \acs{rfm}}&\hspace{2ex}\subfloat[]{\includegraphics[width=0.45\columnwidth]{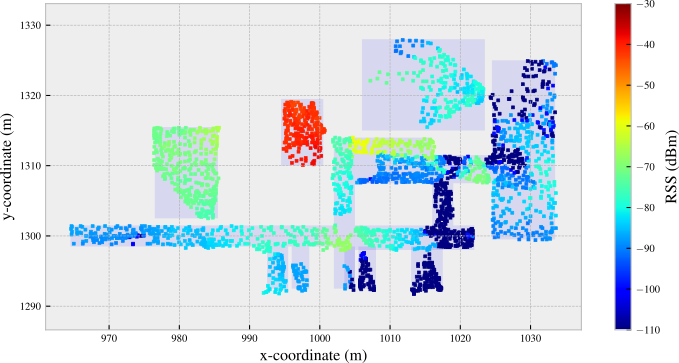}
			\label{subfig:med_rss_1}}&\hspace{2ex}
		\subfloat[]{\includegraphics[width=0.45\columnwidth]{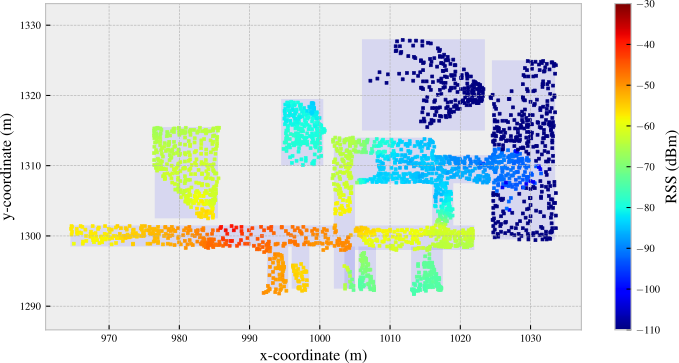}
			\label{subfig:med_rss_2}}\vspace{-1ex}\\
		\rowname{Residual}&\hspace{2ex}\subfloat[]{\includegraphics[width=0.45\columnwidth]{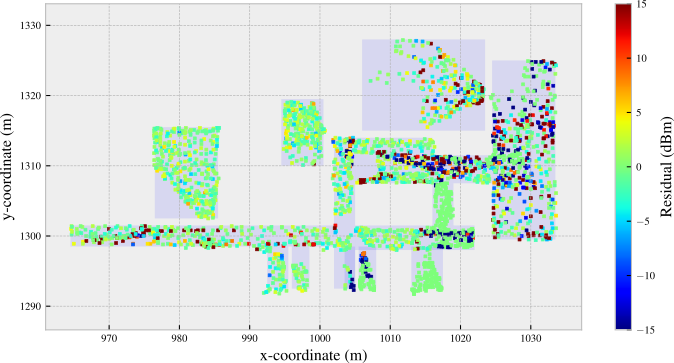}
			\label{subfig:med_res_1}}&\hspace{2ex}
		\subfloat[]{\includegraphics[width=0.45\columnwidth]{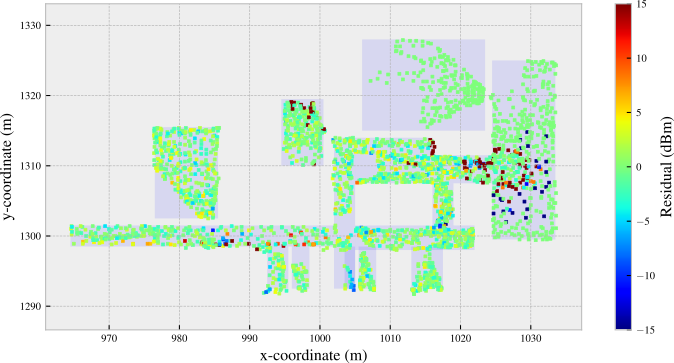}
			\label{subfig:med_res_2}}\vspace{-1ex}
	\end{tabular}
	\caption[Examples of the raw and spatially filtered \acs{rfm} of two arbitrarily selected \acsp{ap}]{Examples of the raw and spatially filtered \acs{rfm} of two arbitrarily selected \acsp{ap}. Third row shows residuals between the spatially filtered \acs{rfm} and the raw \acs{rfm}. The density of the reference locations over the \acs{roi} varies due to different accessibility (\eg areas blocked by furniture or other facilities) and by different visiting frequency of the users.}
	\label{fig:raw_med_rss}
\end{figure}

\figPref\ref{subfig:raw_rss_1} and \ref{subfig:raw_rss_2} show examples of the raw data collected for \acs{rfm} generation, namely the \acs{rss} values from two \acs{wlan} \acsp{ap}. These are signals of opportunity as the \acsp{ap} had been installed for providing Internet access and the signals are their anyway, when using them for the purpose of indoor positioning. The raw measurements have been acquired at arbitrary locations throughout the \acs{roi} which consists of several rooms and corridors within an office building.

\section{Robust estimation of the feature variability}\label{sec:robust_var}
To estimate the noise of the measurable features at each location throughout the \acs{roi}, the features would have to be measured (ideally consecutively) multiple times at each location. However, even for a relatively sparse set of reference points throughout the \acs{roi} this would be prohibitively time-consuming and labor-intensive. We relax this requirement by assuming that the expected feature values change only little within a local, spatial neighborhood. Therefore, instead of estimating the standard deviation from the data collected only at a single location, we use all feature values obtained within a certain radius about a chosen reference location. The corresponding data are identified within the time series of data resulting while the user walked through the \acs{roi}. We denote these fingerprints as {\it kinematically collected} ones. The estimation of the standard deviation is still possible if a sufficient number of measurements is obtained in the proximity of each reference location (see \figPref\ref{fig:raw_med_rss}). The measurements thus associated with an individual reference location contain data obtained consecutively within a short time at slightly different positions, but also data collected a certain time interval apart (\eg half an hour) because the user passed most locations several times during the entire data collection process. The resulting standard deviations thus reflect also the temporal variability of the signals, and the impact of user motion during measurement, which will also apply during the positioning stage. We thus consider the kinematically collected RFM data suitable for the variability analysis.

Under the assumption that the expected value of each feature is locally obtainable, the location-wise STD of each feature can be approximated based on the measurements associated to the neighborhood of a given reference location. More formally, we estimate the \acs{std} $ \sigma_{jk} $ of $ k $-th ($ k=\intSet{|\setSymScript{\tilde{O}}{j}{}|} $) feature at the reference location $ \vecSym{l}_j $ in the \acs{rfm} $ \setSym{F} $. These estimated values of the \acs{std} are later included in the extended representation of the \acs{rfm}, \ie $ \setSym{F}:= \{\vecSymScript{l}{j}{}, \setSymScript{\tilde{S}}{j}{}\} $ with $ \setSymScript{\tilde{S}}{j}{} = \{(\norSymScript{a}{jk}{},\norSymScript{v}{jk}{},\norSymScript{\sigma}{jk}{})| \norSymScript{a}{jk}{}\in \setSymScript{\tilde{A}}{j}{}\} $. We start the estimation of the feature values for the \acs{rfm} by applying a spatial median filter to the raw measurements in order to mitigate potential outliers. We proceed with the \acf{ks} that enables us to reduce the impact of noise and obtain a quasi-continuous representation of the \acs{rfm} by interpolation. It allows us to approximate the expected value of the measurements at any location throughout the \acs{roi}. We perform spatial filtering and \acs{ks} in two separate steps because \acs{ks} is non-robust and the preceding filtering allows us to remove outliers before filtering noise and interpolating. The location-wise \acs{std} for each measurable feature is finally calculated as empirical standard deviation of the raw measurements (before filtering and kernel smoothing) within a neighborhood of the specific reference points.  In the following, the individual steps of the algorithm are explained in more detail.

\begin{figure}[!t]
	\centering
	\includegraphics[width=0.45\linewidth]{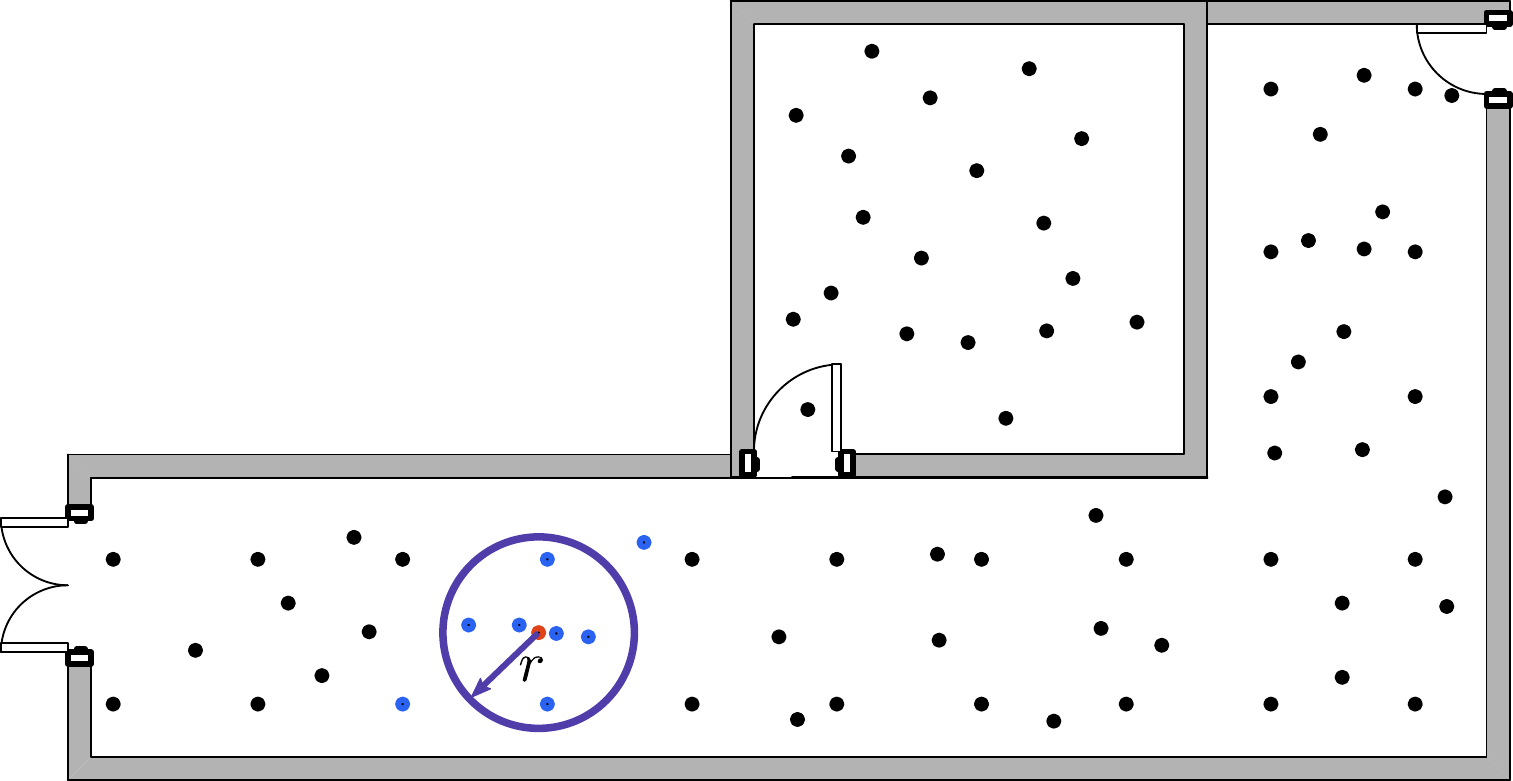}
	\caption[Schematic representation of the spatial filtering]{Schematic representation of the spatial filtering}
	\label{fig:sf}
\end{figure}

As can be seen in \figPref\ref{subfig:raw_rss_1} and \ref{subfig:raw_rss_2} the measured feature values in the neighborhood of a given location may vary significantly. This is particularly visible around locations with very low signal strength values, i.e. values close to the sensitivity limit of the mobile devices. In order to mitigate the impact of these variations on the representation of the \acs{rfm}, we apply the spatial filtering which replaces the originally measured feature value $ v_a $ of feature $ a $ at the given location $ \vecSym{l} $ by the median value of the values measured within the neighborhood of $ \vecSym{l} $. We have chosen to defined the neighborhood as the set of measurements collected at the up to $ m $ locations closest to $ \vecSym{l} $ that at the same time lie within the given radius $ r $ about $ \vecSym{l} $ (see the schematic in \figPref\ref{fig:sf}).

In the second step, we estimate a continuous \ac{rfm} using \ac{ks} in order to be able to retrieve the expected measurements at any location within the \acs{roi} \citep{Berlinet2011}. Albeit \acs{ks} can reduce noise by implicit filtering, it is not robust and the results could therefore be severely contaminated by outliers in the measured features (\figPref\ref{subfig:ks_raw_rss_1} and \ref{subfig:ks_raw_rss_2}). Therefore, we apply \acs{ks} to the media filtered data rather than to the original ones. Because the structure of the indoor region is not taken into account, \acs{ks} tends to smoothen the \acs{rfm} over discontinuities like large changes of feature values or change from feature presence to feature absence over short distances e.g. because of walls. This over-smoothing degrades the quality of the \acs{rfm} for certain features at certain locations. This may be relevant for positioning \citep{10.1007/978-3-642-32645-5_18}, especially when using radio frequency signals such as \acs{wlan} whose propagation is highly influenced by obstacles. Herein we employ a modified version of \acs{ks} which uses only a subset of the data in the neighborhood of a given location for approximating the expected feature values \citep{Berlinet2011}. This alleviates the impact of over-smoothing, while at the same time reducing the computational complexity \citep{Berlinet2011,Cormen:2009:IAT:1614191}.\footnote{A detailed analysis of the over-smoothing problem, the computational complexity of \acs{ks}, and a discontinuity preserving approach to \acs{ks} is beyond the scope of this paper and left for future work.}.

The distribution of the measured noise shown in \figPref\ref{fig:rob_std} clearly suggests that the variances are location-dependent, are different for different features, and cannot be represented as just a function of feature value or of geometric distance from a single point per feature (e.g. the \acs{ap} location). So, we propose to model the \acs{std} as a location-dependent quantity, independently for each individual feature. To this end, we compute the absolute residuals of the raw data with respect to the spatially filtered and kernel smoothed \acs{rfm} in order a robust estimate of the \acs{std}. At the reference location $ \vecSym{l}_j $ in the \acs{rfm} $ \setSym{F} $, the \acs{std} $ \sigma_{jk} $ of $ k $-th ($ k=\intSet{|\setSymScript{\tilde{O}}{j}{}|} $) feature contained in $ \setSym{\tilde{O}}_j $ is computed by the \acf{mad} of the measured feature values associated to locations defined as the support set for spatial filtering. The extended representation of the \acs{rfm} with the estimated \acs{std} at location $ \vecSym{l}_j $ is denoted as $ \setSymScript{\tilde{S}}{j}{} = \{(\norSymScript{a}{jk}{},\norSymScript{v}{jk}{},\norSymScript{\sigma}{jk}{})| \norSymScript{a}{jk}{}\in \setSymScript{\tilde{A}}{j}{}\} $ and is continuously represented using \acs{ks}, \ie $ \setSymScript{\tilde{S}}{j}{} := \funSym{F}(\vecSymScript{l}{j}{}) $.

\section{Iterative scheme for online positioning}\label{sec:iter_pos}
Inspired by the finding that the variability of the features has a large effect on the positioning error  \citep{KAEMARUNGSI2012292}, we employ the robustly estimated \acs{std} of the features to reduce the impact of uncertain feature values when calculating the position estimate. We construct a weighting scheme that reduces the weight of a feature with high \acs{std} relative to features with low \acs{std}. Therefore, a discrepancy between online measured and expected value of a feature with low \acs{std} has more impact on the dissimilarity measure---and thus on the estimated position---than the same discrepancy for a feature with a high \acs{std}. This dissimilarity measure is used to identify which subset of reference locations is taken into account when inferring the user's location using deterministic feature-based positioning algorithms such as \acs{knn}.

\subsection{Weighted dissimilarity measure}\label{subsec:wdm}

Given the online measured features $ \setSymScript{O}{i}{u} $ at the location $ \vecSymScript{l}{i}{u} $, the weighted dissimilarity measure $ \norSymScript{\funSym{d}}{}{w} $ between $ \setSymScript{O}{i}{u} $ and the $ j $-th reference fingerprint $ \setSymScript{\tilde{O}}{j}{} $ stored in the \acs{rfm} is computed as:
\begin{equation}
\label{eq:wcdm}
\norSymScript{\funSym{d}}{}{w}(\setSymScript{O}{i}{u}, \setSymScript{\tilde{O}}{j}{}) =  \sum\limits_{a\in \setSymScript{A}{i}{u}\cap\setSymScript{\tilde{A}}{j}{}}\norSymScript{w}{ik}{u}\,\funSym{g}(\norSymScript{v}{ik}{u}, \norSymScript{v}{jk}{}) + \norSymScript{\alpha}{1}{} \cdot\sum\limits_{a\in \setSymScript{A}{i}{u}\backslash\setSymScript{\tilde{A}}{j}{}}\norSymScript{w}{ik}{u}\,\funSym{g}(\norSymScript{v}{ik}{u}, \gamma) + \norSymScript{\alpha}{2}{} \cdot\sum\limits_{a\in \setSymScript{\tilde{A}}{j}{}\backslash\setSymScript{{A}}{i}{u}}\norSymScript{w}{ik}{u}\,\funSym{g}(\gamma, \norSymScript{v}{jk}{})
\end{equation}
where $ \funSym{g} $ is the selected dissimilarity measure (\eg Minkowski distance) and $ \gamma $ is the missing value indicator (\eg -110~dBm). This equation represents a \acf{cdm} as defined in \citep{Zhou2018ipin}, and correspondingly $ \norSymScript{\alpha}{1}{} $ and $ \norSymScript{\alpha}{2}{} $ are hyperparameters regulating the contribution of mutually unshared features to the dissimilarity measure. However, the \acs{cdm} herein uses a new distance metric, not covered in \citep{Zhou2018ipin}, by location-wise weighting of individual features instead of only weighting according to the respective observability. $ \norSymScript{w}{ik}{u} $ is the weight of the $ k $-th feature at the location/time $ i $ and is computed by employing the variability derived from the estimated expected measurement $ \setSymScript{\tilde{S}}{i}{u} $ obtained at $ \vecSymScript{l}{i}{u} $. In case that the $ k $-th feature in $ \setSymScript{\tilde{A}}{ij}{} $ ($ \setSymScript{\tilde{A}}{ij}{} := \setSymScript{{A}}{i}{u} \cup \setSymScript{\tilde{A}}{j}{} $) is not measurable at location $ \vecSymScript{l}{i}{u} $, the weight of the corresponding feature is set to the minimum value of the weights of the measurable features thus reducing their impact on the estimation of the location.

We selected the softmax function \citep{Murphy:2012:MLP:2380985,gal2016dropout}
\begin{equation}
\label{eq:w_softmax}
\norSymScript{w}{ik}{u} = \frac{{\operatorname{e}}^{-\beta \norSymScript{\sigma}{ik}{-2}}}{\sum\limits_{l=1}^{|\setSymScript{\tilde{S}}{i}{u}|}{{\operatorname{e}}^{-\beta \norSymScript{\sigma}{lk}{-2}}}}
\end{equation}
to calculate the weight of each feature using the estimated \acs{std}, where $ \setSymScript{\tilde{S}}{i}{u} = \funSym{F}(\vecSymScript{l}{i}{u}) $ and $ \beta > 0 $ is the scale factor for adapting the concentration of the softmax function. The denominator normalizes the weights and makes the solution invariant to the scale of the weights. We have also tried to use a weight function corresponding to the one frequently employed for weighted least-squares (and actually motivated by maximum likelihood estimation with normally distributed observations), namely setting each weight proportional to the inverse of the respective variance. However, the accuracy of the solutions was worse than using the softmax function.

The weighted dissimilarity measure is used to identify the candidate locations, whose dissimilarity values are smallest among all reference fingerprints stored in the \acs{rfm}. We estimate the user's location using \acs{knn} or weighted \acs{knn} by averaging or weighted averaging (\eg inversely proportional to the value of dissimilarity measure) of the candidate locations. More details about \acs{knn} and weighted \acs{knn} can be found in \eg \citep{Padmanabhan2000, Zhou2018}.

\subsection{Iterative scheme}\label{subsec:iter_sch}
The position estimation requires to calculate the weight of each feature. However, the weight depends on the standard deviation which in turn varies with location. The required value can only be extracted from the \acs{rfm} once the location is known. We thus carry out the positioning in an iterative way by i) assuming a position (initialization); ii) retrieving the \acsp{std} from the \acs{rfm}, calculating the weights and estimating the position (update step); and iii) repeating ii) until a termination condition of the iterative scheme is fulfilled. These steps are explained in more detail in the following subsections.

\subsubsection{Determination of the initial location}
The initial location $ \vecSymScript{\hat{l}}{i}{(0)} $ of the user is used to derive the weights for the first iteration. One straightforward way of initializing is to choose the location estimated by the standard \acs{knn} without the weighted dissimilarity measure (\ie the traditional \acs{knn}). When processing real world data we found out that the solution obtained at the termination of the iterative process is quite stable when initializing the location even randomly (see \secPref\ref{sec:perf_ana}). This suggests that the positioning performance does not depend strongly on the choice of the initial location.

\subsubsection{Update step}
At the $ t $-th iteration ($ t \in\convSetSym{N}^{+} $), the weights as well as the dissimilarities are computed according to the variability obtained at the location searched at the $ (t-1) $-th iteration. The weight $ \norSymScript{w}{ik}{(t)} $ of the $ k $-th feature at location/time $ i $ and the $ t $-th iteration is defined as: 
\begin{equation}
\label{eq:w_update}
\norSymScript{w}{ik}{(t)} = \frac{{\operatorname{e}}^{-\beta \big({\norSymScript{\sigma}{lk}{(t-1)}}\big)^{-2}}}{\sum\limits_{l=1}^{|\setSymScript{\tilde{S}}{i}{(t-1)}|}{{\operatorname{e}}^{-\beta \big({\norSymScript{\sigma}{lk}{(t-1)}}\big)^{-2}}}}
\end{equation}
where $ \setSymScript{\tilde{S}}{i}{(t-1)} $ is the estimated expected value of features with their \acs{std} at location $ \vecSymScript{\hat{l}}{}{(t-1)} $. This updated weights are used to compute the dissimilarity measure as defined in \eqref{eq:wcdm} and consequently to infer the estimated location $ \vecSymScript{\hat{l}}{i}{(t)} $ at the $ t $-th iteration using \eg \acs{knn} algorithm.

\subsubsection{Termination condition}
Ideally the searching process should converge to a fixed location. This state is assumed to be reached when the distance between two consecutively obtained location estimated is lower than a given small threshold. We denote this subsequently as converging state and terminate the iterative process when
\begin{equation}
\label{eq:term_conv}
|\vecSymScript{\hat{l}}{i}{(t)} - \vecSymScript{\hat{l}}{i}{(t-1)}|_2 < \norSymScript{d}{}{min}, \nonumber
\end{equation}
where $ \norSymScript{d}{}{min} $ is the threshold, which we set to $ 10^{-3} $~m in the experimental analysis later on. We found out that the iterative process proposed herein sometimes enters a loop in which a (small) subset of locations are repeatedly obtained as estimates in the same sequence. We denote this as the looping state and introduce a second termination condition which is met when this state is recognized. We implement it as a threshold on the distance between the location estimate obtained at the iteration $ t $ and the ones estimated at previous iterations except the estimated location at the $ (t -1) $-th iteration. More formally, the second condition is satisfied and the iteration is terminated when
\begin{equation}
\label{eq:term_loop}
\underset{m=1,\cdots, t - 2}{\operatorname{min}}\{|\vecSymScript{\hat{l}}{i}{(t)} - \vecSymScript{\hat{l}}{i}{(m)}|_2\} < \norSymScript{d}{}{min}. \nonumber
\end{equation}

Finally, the maximum number $ T $ of iterations is also limited (\eg $ T=100 $) in order to prevent long or endless search for a solution. If the search for an estimate is terminated due to this condition, we denote it as max.~state. 


Assuming that the iterations terminate after $ T' $ iterations we select or compute the final estimate of the position depending on the \acf{tf} $ \norSymScript{\epsilon}{i}{u} \in\{0, 1, 2\} $, indicating the respective state, as follows:
\begin{itemize}
	\item \textbf{Converging state}: The location estimated at the $ T' $-th iteration is selected as the final estimate of the user's location $ \vecSymScript{\hat{l}}{i}{u} $, and $ \norSymScript{\epsilon}{i}{u} $ is set to 0.
	\item \textbf{Looping state}: In this case the searched locations do not converge to a single point. If the number of locations exceeds a certain minimum (\eg 4) and if the locations visited in the looping state are not farther apart than a chosen maximum (\eg 0.01~m) (see \figPref\ref{fig:iter_pos}) we use the \acf{mcd} estimator\footnote{\acs{mcd} is a highly robust estimation of multivariate location and scatter. We use the implementation of \acs{mcd} from scikit-learn \citep{scikit-learn}.} for computing the estimated location $ \vecSymScript{\hat{l}}{i}{u} $ of the user from the  convex hull of the visited locations. If the number of points is too low or if they are too far apart from each other the situation is handled like the max.~state. If the looping state termination condition is met and the \acs{mcd} is reported as the final estimate, the \acs{tf} $ \norSymScript{\epsilon}{i}{u} $ is set to 1. 
	\item \textbf{Max.~state}: This case actually means that the position estimation using the weighted dissimilarity measure fails because no position can be found where the measured features and the predetermined standard deviations are compatible. In this case, we can either report a failure of the algorithm and not calculate a solution, or we can calculate an estimate ignoring the variability information. We have chosen the latter herein. In particular, we determine $ \vecSymScript{\hat{l}}{i}{u} $ from all searched locations $ \setSymScript{\hat{L}}{}{} $ analyzing the similarities between the user measured fingerprint $ \setSymScript{O}{i}{u}  $ and the expected ones at the searched locations. Specifically, we employ the \acf{mji}, which has been used for identifying subregions according to the measurability of features \citep{Zhou2018}, as the similarity metric. The \acs{mji} value $ \funSym{S}^{\mathrm{MJI}} $ between $ \setSymScript{O}{i}{u}  $ and the expected one $ \setSymScript{\tilde{\hat{O}}}{i}{(t)} (\setSymScript{\tilde{\hat{O}}}{i}{(t)}:= \funSym{F}(\vecSymScript{\hat{l}}{i}{(t)})) $ at the searched location $ \vecSymScript{\hat{l}}{i}{(t)} $ of the $ t $-th iteration is computed by: 
	\begin{equation}
	\label{eq:mji}
	\funSym{S}^{\mathrm{MJI}}(\setSymScript{O}{i}{u}, \setSymScript{\tilde{\hat{O}}}{i}{(t)}) = \frac{1}{2}\Big(\frac{|\setSymScript{A}{i}{u}\cap \setSymScript{\tilde{\hat{A}}}{i}{(t)}|}{|\setSymScript{A}{i}{u}\cup \setSymScript{\tilde{\hat{A}}}{i}{(t)}|} + \frac{|\setSymScript{A}{i}{u}\cap \setSymScript{\tilde{\hat{A}}}{i}{(t)}|}{|\setSymScript{A}{i}{u}|}\Big)
	\end{equation}
	where $ \setSymScript{A}{i}{u} $ and $ \setSymScript{\tilde{\hat{A}}}{i}{(t)} $ are the sets of the measured features contained in $  \setSymScript{O}{i}{u} $ and $ \setSymScript{\tilde{\hat{O}}}{i}{(t)} $, respectively. The estimated user's location $ \vecSymScript{\hat{l}}{i}{u} $ is then the one that has the biggest \acs{mji} value among all searched locations $ \setSymScript{\hat{L}}{}{} $, \ie the one with the maximum number of common measurable features is selected as the final estimate of the user's location.
\end{itemize}

\section{Analysis of the variability estimation and positioning performance}\label{sec:perf_ana}

We start this section by presenting the results of the location-wise variability of each individual feature estimated using the kinematically collected \acs{rfm} data, which is discussed in detail in \citep{Zhou2018}. We conclude this section with an analysis concerning the characteristics of iteratively searched locations as well as the positioning performance of the proposed iterative scheme.

\subsection{Results of the variability estimation}
Herein we set $m= 20$ and $r=2$~m to obtain the spatially filtered \acs{rfm}, which visually has an adequate spatial consistency in the neighborhood of each location. \figPref\ref{subfig:med_rss_1} and \ref{subfig:med_rss_2} clearly show that the spatial filtering can reduce the large variations contained in the raw \acs{rfm} to a great extent. For further analysis, we compute the residuals between the raw and the spatially filtered \acs{rfm}. The obtained residuals are close to zero-mean distributed and have a location-dependent magnitude as illustrated in \figPref\ref{subfig:med_res_1} and \ref{subfig:med_res_2}. Large residuals occur either in regions close to the boundaries of the \acs{roi} (\eg close to the walls or corners of rooms and corridors) or at locations where the \acs{rss} values are hardly measurable by the mobile phone. In both cases the features are very likely affected by obstacles which also cause locally large variations of the feature values.
\begin{figure}[!h]
	\centering
	\settoheight{\tempdima}{\includegraphics[width=.45\linewidth]{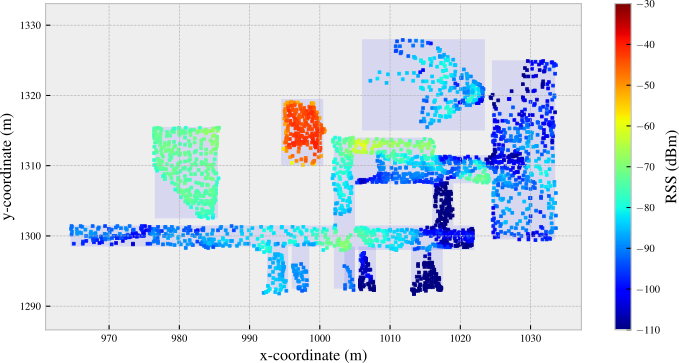}}
	\begin{tabular}{c@{ }c@{ }c@{ }}
		&\acs{ap}: \textit{9c:50:ee:09:5f:30} & \acs{ap}: \textit{9c:50:ee:09:61:d1}\vspace{-1.5ex}\\
		\rowname{\shortstack{\acs{ks} using \\raw \acs{rfm}}}&\hspace{2ex}\subfloat[]{\includegraphics[width=0.45\columnwidth]{rss_ks_mac_9c_50_ee_09_5f_30.png}
			\label{subfig:ks_raw_rss_1}}&\hspace{1.5ex}
		\subfloat[]{\includegraphics[width=0.45\columnwidth]{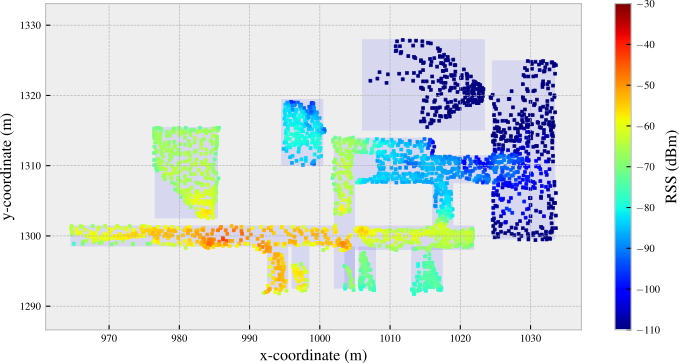}
			\label{subfig:ks_raw_rss_2}}\vspace{1ex}\\
		\rowname{\shortstack{\acs{ks} using \\spatially filtered \acs{rfm}}}&\hspace{2ex}\subfloat[]{\includegraphics[width=0.45\columnwidth]{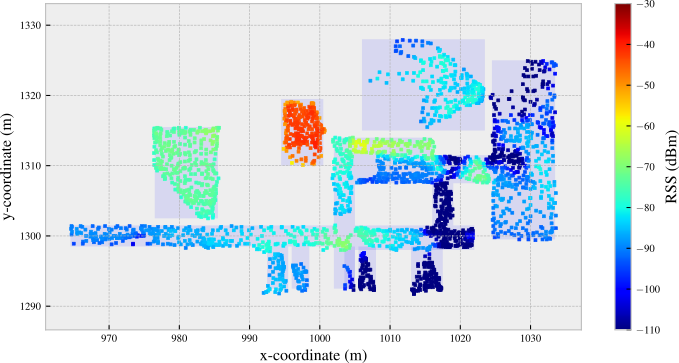}
			\label{subfig:ks_med_rss_1}}&\hspace{1.5ex}
		\subfloat[]{\includegraphics[width=0.45\columnwidth]{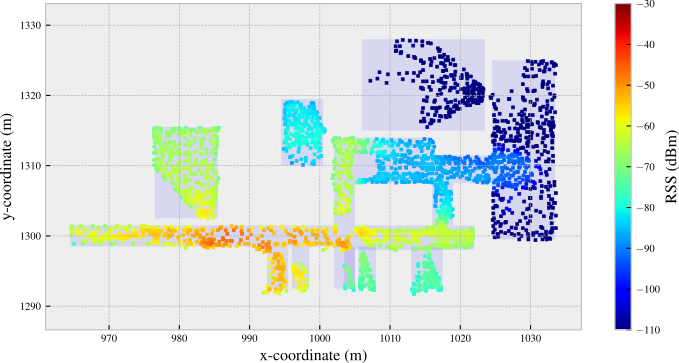}
			\label{subfig:ks_med_rss_2}}\vspace{1ex}\\
		\rowname{Residual}&\hspace{2ex}\subfloat[]{\includegraphics[width=0.45\columnwidth]{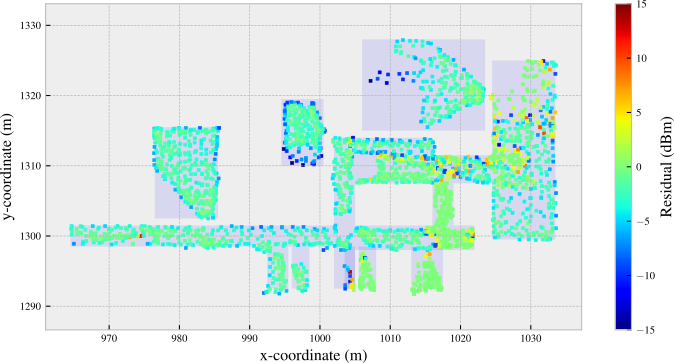}
			\label{subfig:ks_med_res_1}}&\hspace{1.5ex}
		\subfloat[]{\includegraphics[width=0.45\columnwidth]{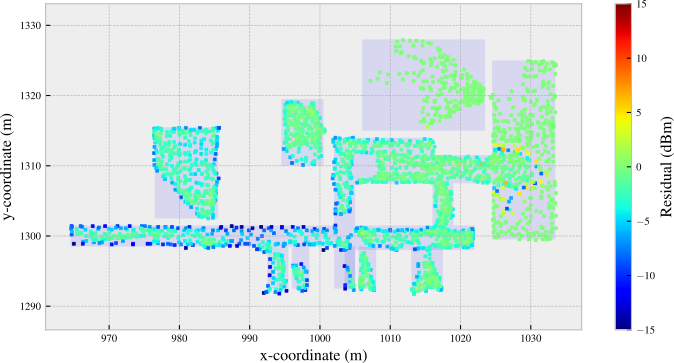}
			\label{subfig:ks_med_res_2}}\vspace{-1ex}
	\end{tabular}
	\caption[Examples of kernel smoothed \acs{rfm} of two arbitrary selected \acsp{ap}]{Examples of kernel smoothed \acs{rfm} of two arbitrary selected \acsp{ap}. The results of the first two rows are yielded by taking the raw \acs{rfm} and the spatially filtered one (\ie the ones depicted in the first two rows of \figPref\ref{fig:raw_med_rss}) as the input to \acs{ks}, respectively. Third row shows the residual between the kernel smoothed \acs{rfm} using the spatially filtered one and the spatially filtered \acs{rfm}. Though the \acs{ks} provides the continuous representation of the \acs{rfm}, we only visualize the smoothed features at the locations contained in the raw \acs{rfm} for easy comparison.}
	\label{fig:ks_raw_med_rss}
\end{figure}

\begin{figure}[!h]
	\centering
	\settoheight{\tempdima}{\includegraphics[width=.45\linewidth]{rss_ks_mac_9c_50_ee_09_5f_30.png}}
	\begin{tabular}{c@{ }c@{ }c@{ }}
		&\acs{ap}: \textit{9c:50:ee:09:5f:30} & \acs{ap}: \textit{9c:50:ee:09:61:d1}\vspace{-1.5ex}\\
		\rowname{Absolute residual}&\hspace{2ex}\subfloat[]{\includegraphics[width=0.45\columnwidth]{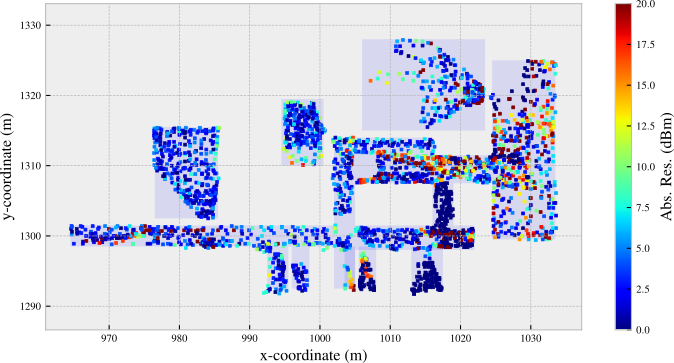}
			\label{subfig:abs_res_1}}&\hspace{1.5ex}
		\subfloat[]{\includegraphics[width=0.45\columnwidth]{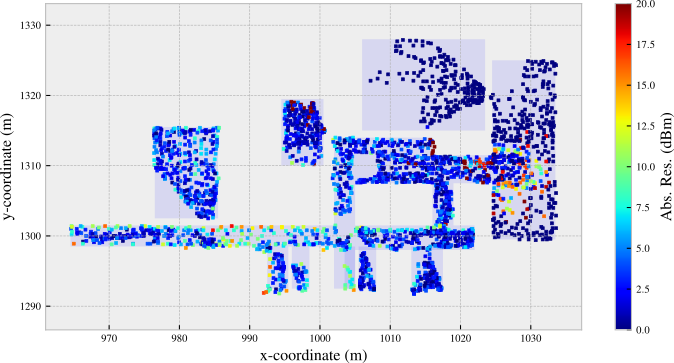}
			\label{subfig:abs_res_2}}\vspace{-1ex}\\
		\rowname{Estimated \acs{std}}&\hspace{2ex}\subfloat[]{\includegraphics[width=0.45\columnwidth]{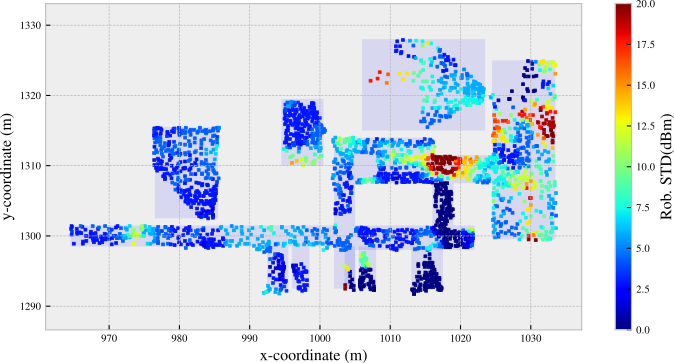}
			\label{subfig:abs_std_1}}&\hspace{1.5ex} 
		\subfloat[]{\includegraphics[width=0.45\columnwidth]{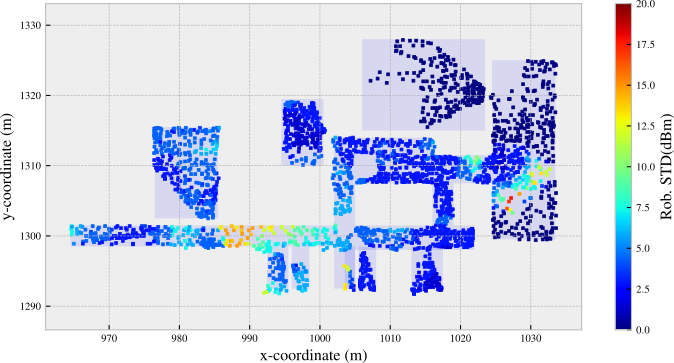}
			\label{subfig:abs_std_2}} 
	\end{tabular}
	\caption{Examples of the absolute residuals and estimated \acs{std} for two \acsp{ap}}
	\label{fig:rob_std}
\end{figure}

\figPref\ref{fig:rob_std} shows the results of the estimated \acs{std} value using the \acs{mad} of the measured feature values associated to the neighborhood of a given location. As can be seen, each feature has a different variability throughout the \acs{roi}, \ie the \acs{std} value is dependent both on the feature as well as on the location. This is the primary motivation that the variability is modeled location-wise for each individual feature instead of simply expressing the variability as a function of the measured feature value or as a constant value. The regions where the feature values have a higher \acs{std} are clearly correlated to the local variations of the measured feature value and the geometry of the building (\figPref\ref{fig:ks_raw_med_rss} and \ref{fig:rob_std}). The high variances occur in the case that a low number of measurements has been collected in the neighborhood region. These are caused by the violation of the assumption that the expected feature values are locally obtainable.

\subsection{Results of iterative scheme for positioning}
The proposed iterative scheme for \acl{fbp} is implemented using the application programming interface of scikit-learn package, a widely used machine learning package in Python \citep{scikit-learn}. Herein we present the results of the iterative positioning using \acs{knn} with the weighted Euclidean distance as the dissimilarity metric for measuring the distance in the feature space. The values of several hyperparameters have to be configured. Regarding the weighted dissimilarity measure, as formulated in \secPref\ref{subsec:wdm}, $ \norSymScript{\alpha}{1}{} $ and $ \norSymScript{\alpha}{2}{} $ are set to 3.0 by following the results reported in \citep{Zhou2018ipin}. The number of nearest neighbors $ k $ in the \acs{knn} algorithm and the scale factor $ \beta $ of the softmax function are empirically set to 1 and 2.0, respectively. Optimization of the parameters (\eg using grid/random search or Bayesian optimization \citep{Wang2014TheoreticalAO}) for achieving the best positioning performance is left for future work.

\begin{figure}[!h]
	\centering
	\settoheight{\tempdima}{\includegraphics[width=0.31\columnwidth]{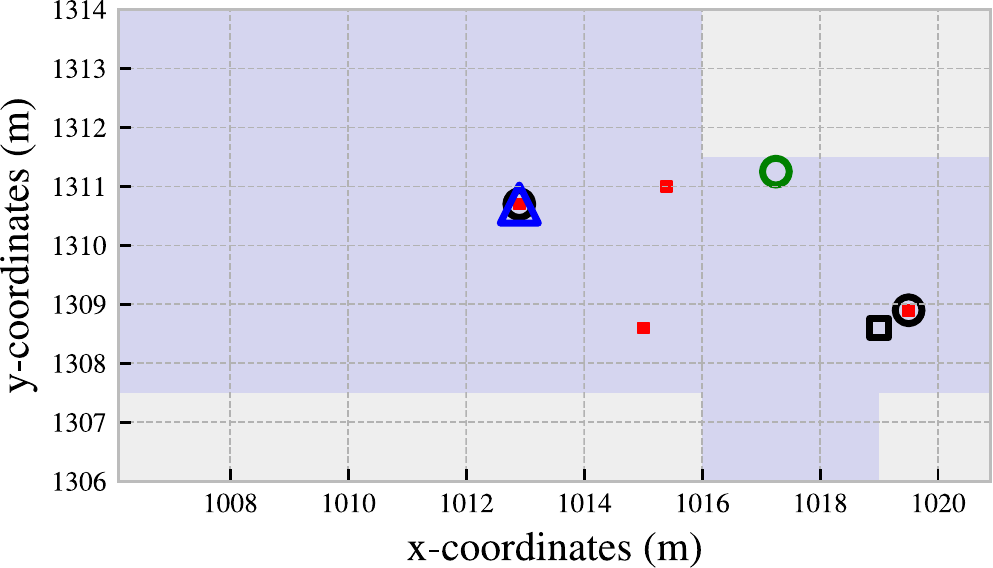}}
	\begin{tabular}{c@{  }c@{ }c@{ }c@{ }}
		& \acs{knn} initialization &Random initialization 1 &Random initialization 2 \vspace{-1.5ex}\\
		\rowname{Loc. 1}&\hspace{0.2ex}
		\subfloat[]{\includegraphics[width=0.31\columnwidth]{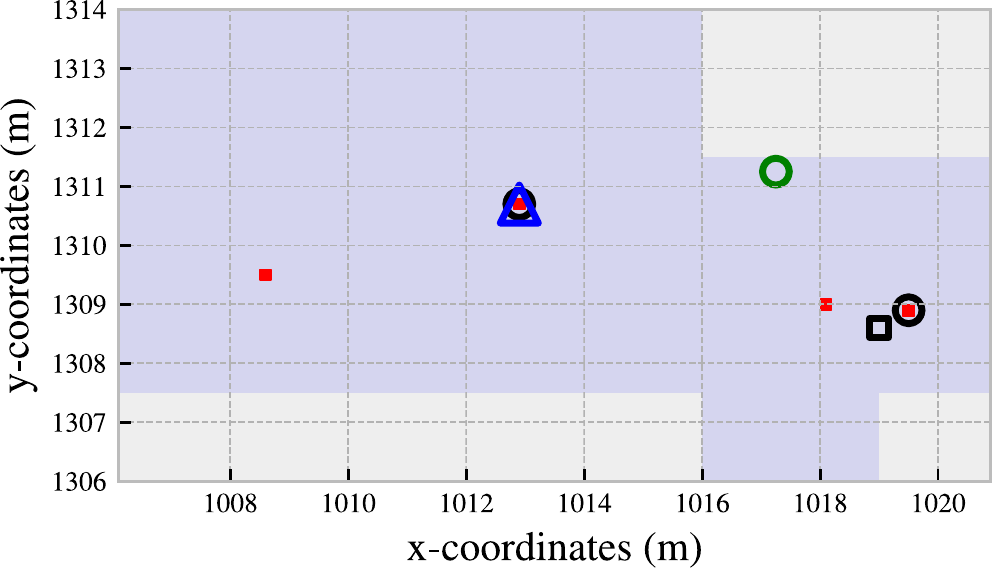}
			\label{subfig:iter_pos_le_0_fix}}\hspace{.2ex} &\subfloat[]{\includegraphics[width=0.31\columnwidth]{tp_625_std_knn_random_2171_iter_pos_wcdm.pdf}
			\label{subfig:iter_pos_le_0_ran_1}}&\hspace{0.2ex}
		\subfloat[]{\includegraphics[width=0.31\columnwidth]{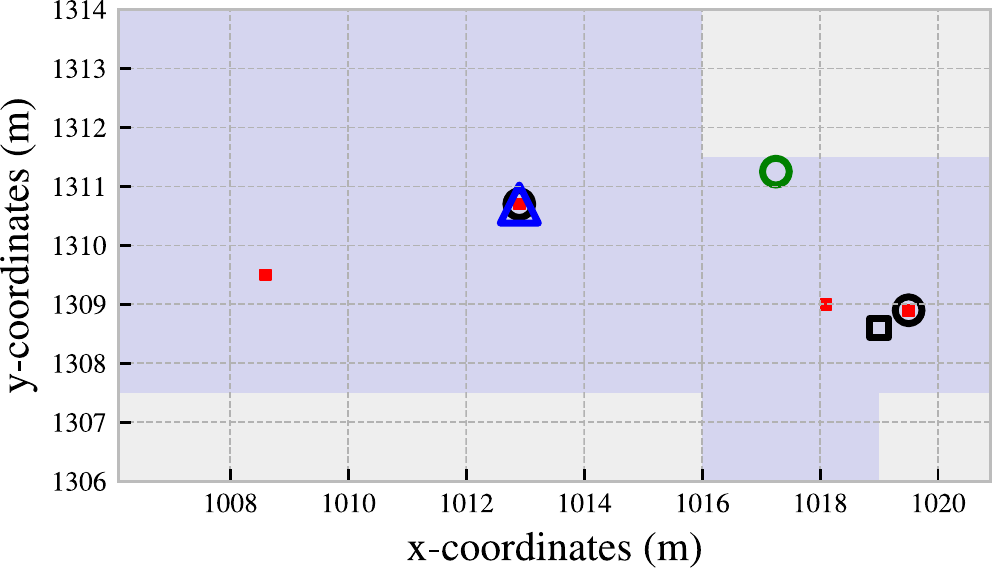}
			\label{subfig:iter_pos_le_0_ran_2}}\vspace{-1ex}\\
		\rowname{Loc. 2}&\subfloat[]{\includegraphics[width=0.31\columnwidth]{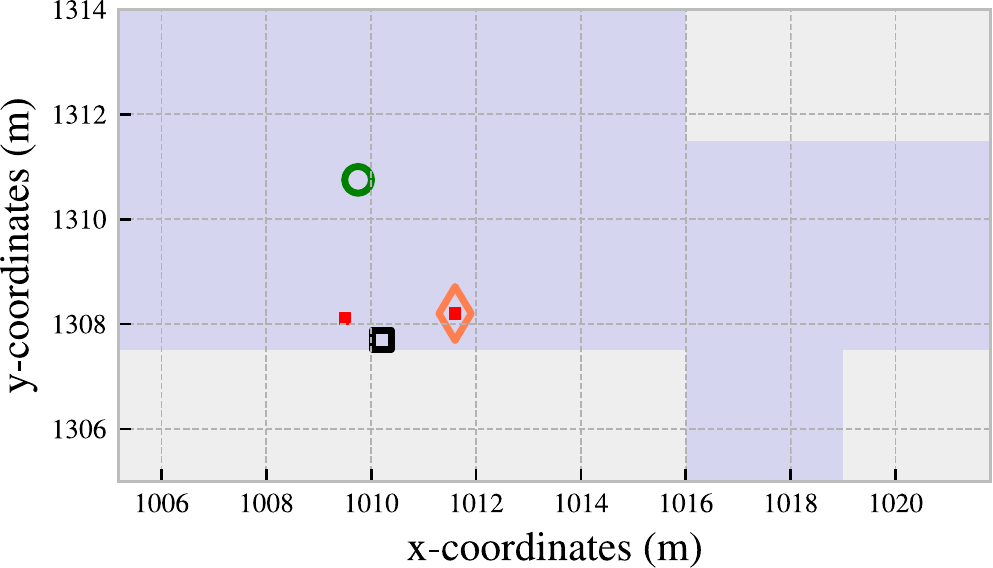}
			\label{subfig:iter_pos_le_929_fix}} &\hspace{.2ex}\subfloat[]{\includegraphics[width=0.31\columnwidth]{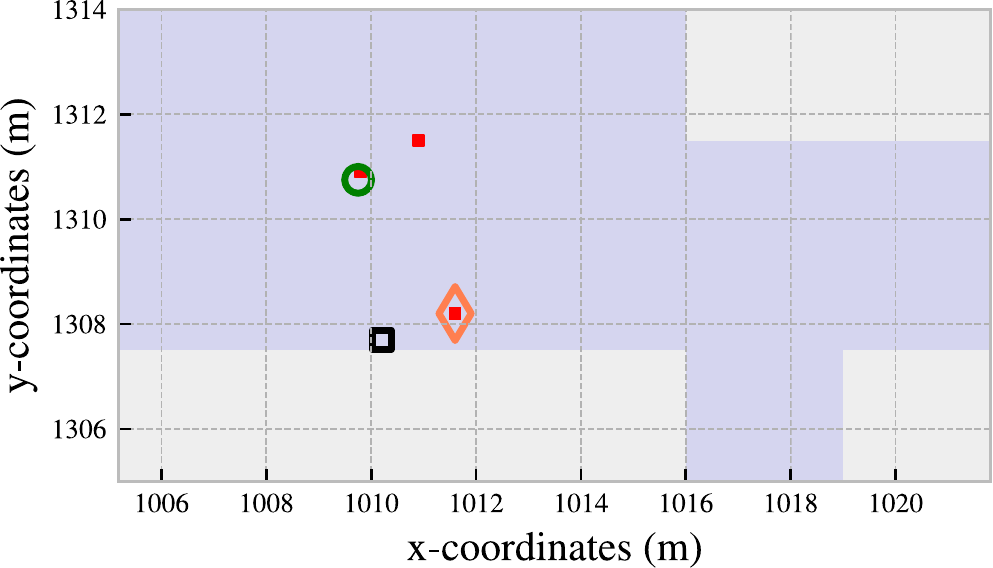}
			\label{subfig:iter_pos_le_929_ran_1}}&\hspace{0.2ex}
		\subfloat[]{\includegraphics[width=0.31\columnwidth]{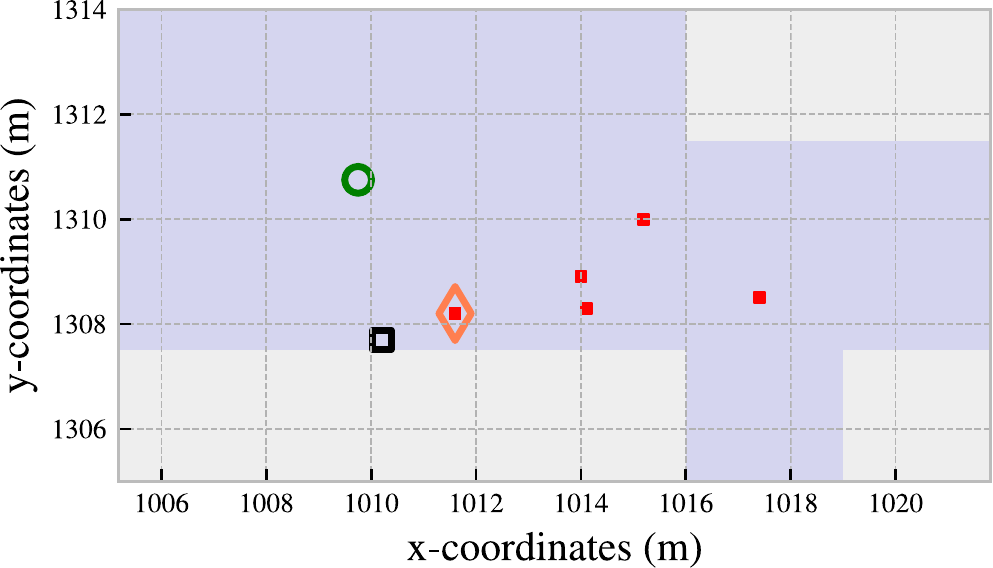}
			\label{subfig:iter_pos_le_929_ran_2}}\hspace{0.2ex}
		\vspace{-1ex}\\
		\rowname{Loc. 3}&\subfloat[]{\includegraphics[width=0.31\columnwidth]{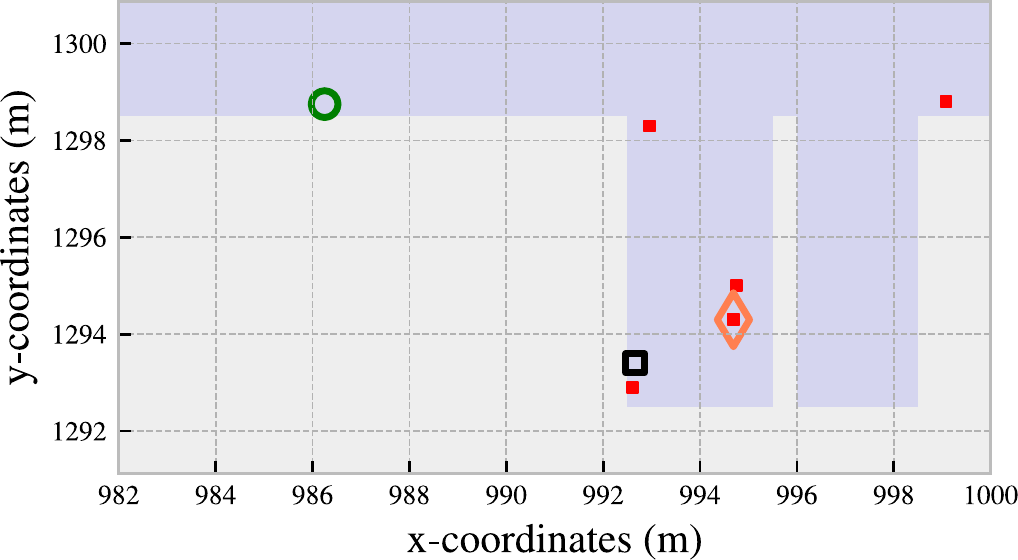}
			\label{subfig:iter_pos_le_581_fix}} &\hspace{.2ex}\subfloat[]{\includegraphics[width=0.31\columnwidth]{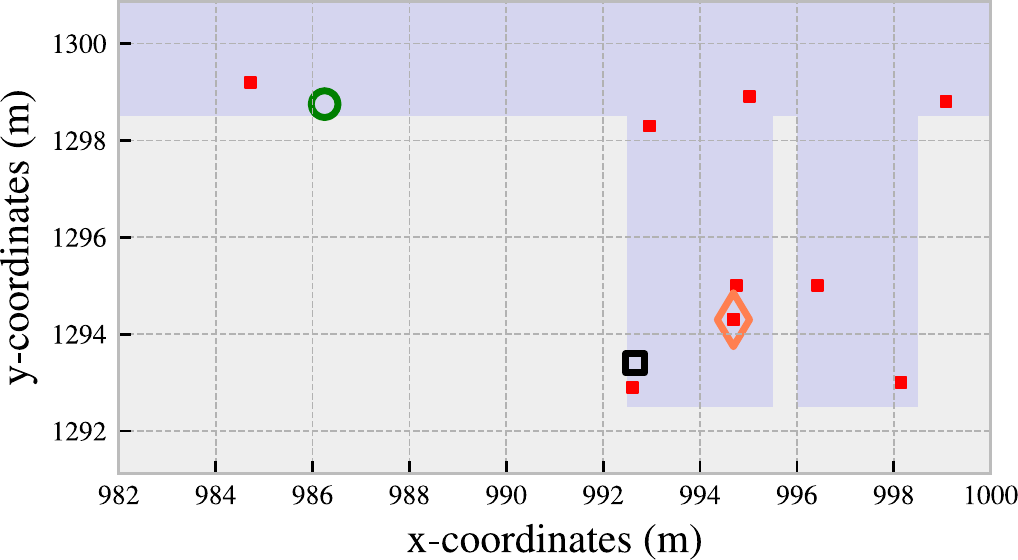}
			\label{subfig:iter_pos_le_581_ran_1}}&\hspace{0.2ex}
		\subfloat[]{\includegraphics[width=0.31\columnwidth]{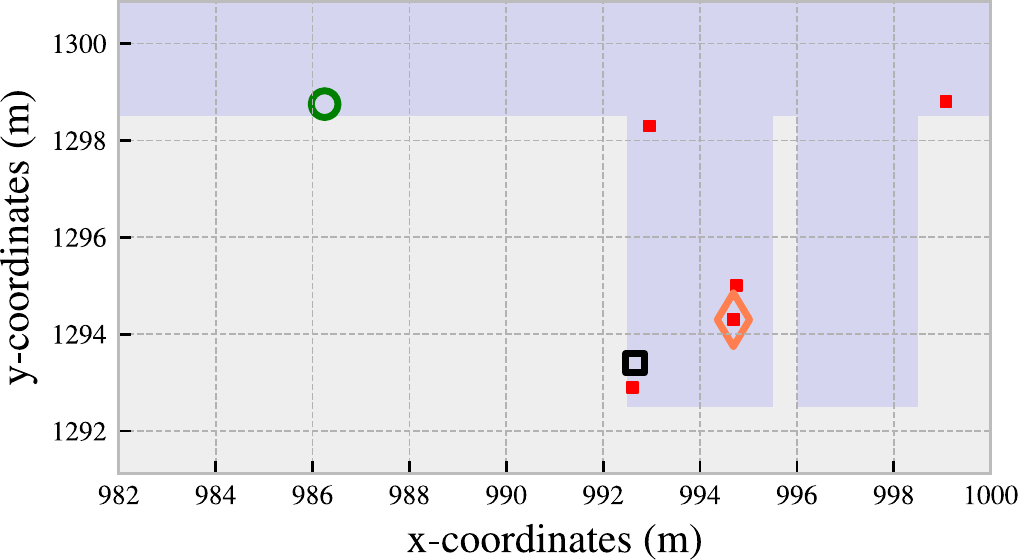}
			\label{subfig:iter_pos_le_581_ran_2}}
		\vspace{1ex}\\
		&\multicolumn{3}{c}{\includegraphics[width=0.8\linewidth]{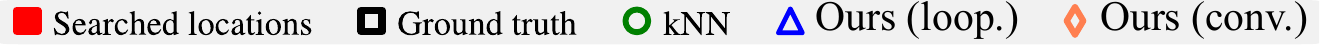}}
	\end{tabular}
	\caption[Examples of iteratively searched location with different initializations]{Examples of iteratively searched location with different initializations. The locations with black circles are repeatedly searched in the same sequence (\ie the looping state) when using the iterative scheme.}
	\label{fig:iter_pos}
\end{figure}

\begin{figure}[!htb]
	\centering
	\includegraphics[width=0.35\linewidth]{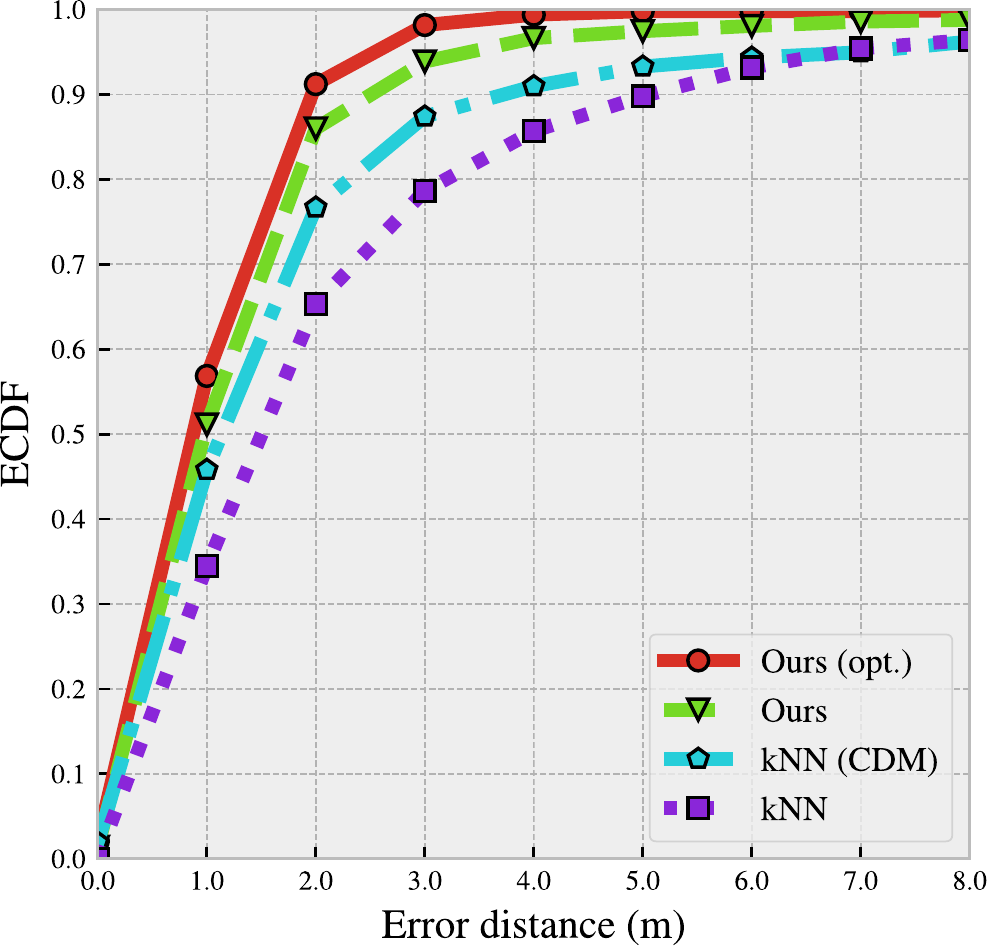}
	\caption[Comparison of \acs{ecdf} \acl{wrt} the radial positioning errors]{Comparison of \acs{ecdf} \acl{wrt} the radial positioning errors.}
	\label{fig:cmp_ecdf_opt}
\end{figure}

\figPref\ref{fig:iter_pos} shows several examples of the searched locations of the iterative positioning with random as well as \acs{knn} initialization. Each individual subplot depicts the results of the iterative positioning at a fixed test location. The subplots depict the searched locations (red squares), the final estimation (blue triangle or coral diamond), the estimation using the traditional \acs{knn} algorithm, and the ground truth (black square)\footnote{{In order to improve the readability, the initial location has not been visualized.}}. The initialization of the initial location has only a minor impact on the iterative searching process in this case because the \acs{roi} is relatively small. The initial location is determined by arbitrarily taking one of the reference locations stored in the \acs{rfm} in case of random initialization. In addition, our schemes for determining the final estimation of the user's location from these searched locations do not achieve the best potential positioning performance using the iterative scheme. Because there are locations in $ \setSym{\hat{L}} $ that are closer to the ground truth but they are not taken as the final estimation. This suggests that the iterative scheme for positioning has the potential to further improve the positioning performance if the proper technique is applied to retrieve the final estimation from these searched locations. The optimal positioning accuracy (denoted as $ \mbox{Ours (opt.)} $ in \figPref\ref{fig:cmp_ecdf_opt} and in \tabPref\ref{tab:stats_pe}) is defined by assuming that the technique for the final estimation is capable of retrieving the searched location that is the closest to the ground truth. As depicted in \figPref\ref{fig:cmp_ecdf_opt}, our scheme for retrieving the final estimation can achieve comparable performance to that of the optimal one when comparing the overall positioning accuracy. However, from \tabPref\ref{tab:stats_pe}, it also suggests that the maximum positioning error can be reduced to a large extent if the optimal positioning can be achieved.

\begin{figure}[!htb]
	\centering
	\includegraphics[width=0.4\linewidth]{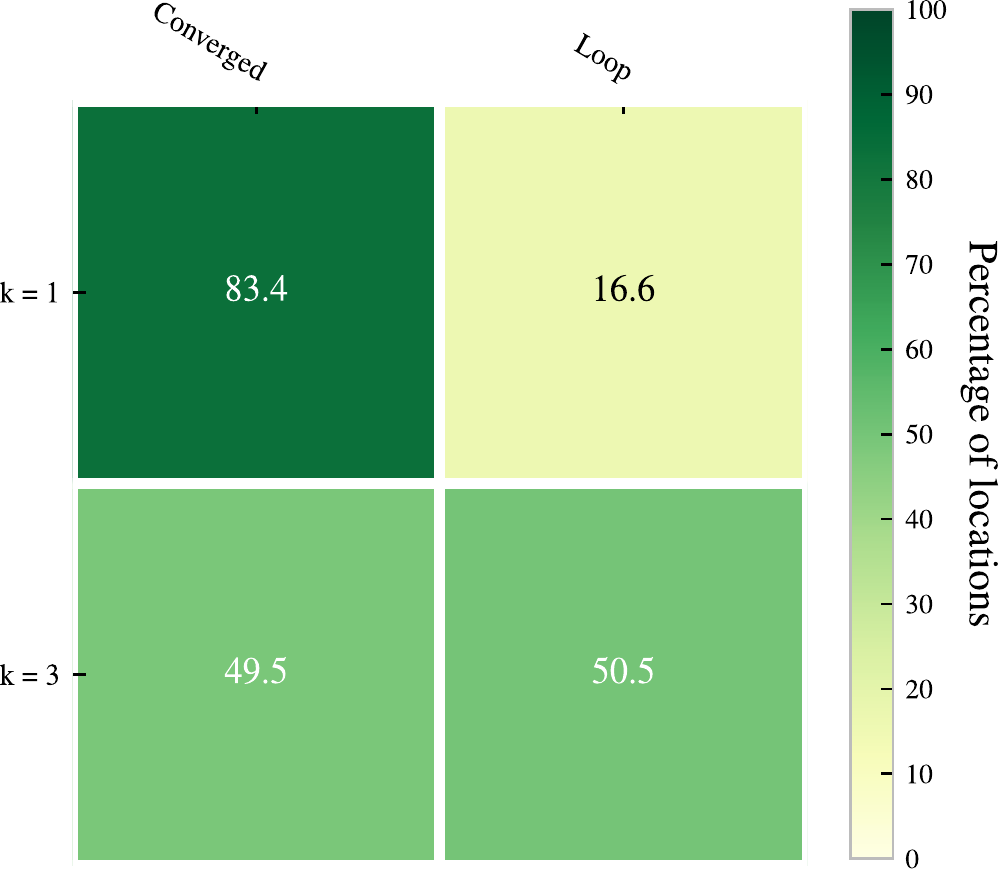}
	\caption[Comparison of the percentage of locations terminated with different conditions for $ k=1\mbox{ and } 3$]{Comparison of the percentage of locations terminated with different conditions for $ k=1\mbox{ and } 3$. The max. state is not included in the figure because it has not happened in the experimental analysis.}
	\label{fig:cmp_tf}
\end{figure}

\begin{figure}[!htb]
	\centering
	\includegraphics[width=0.4\linewidth]{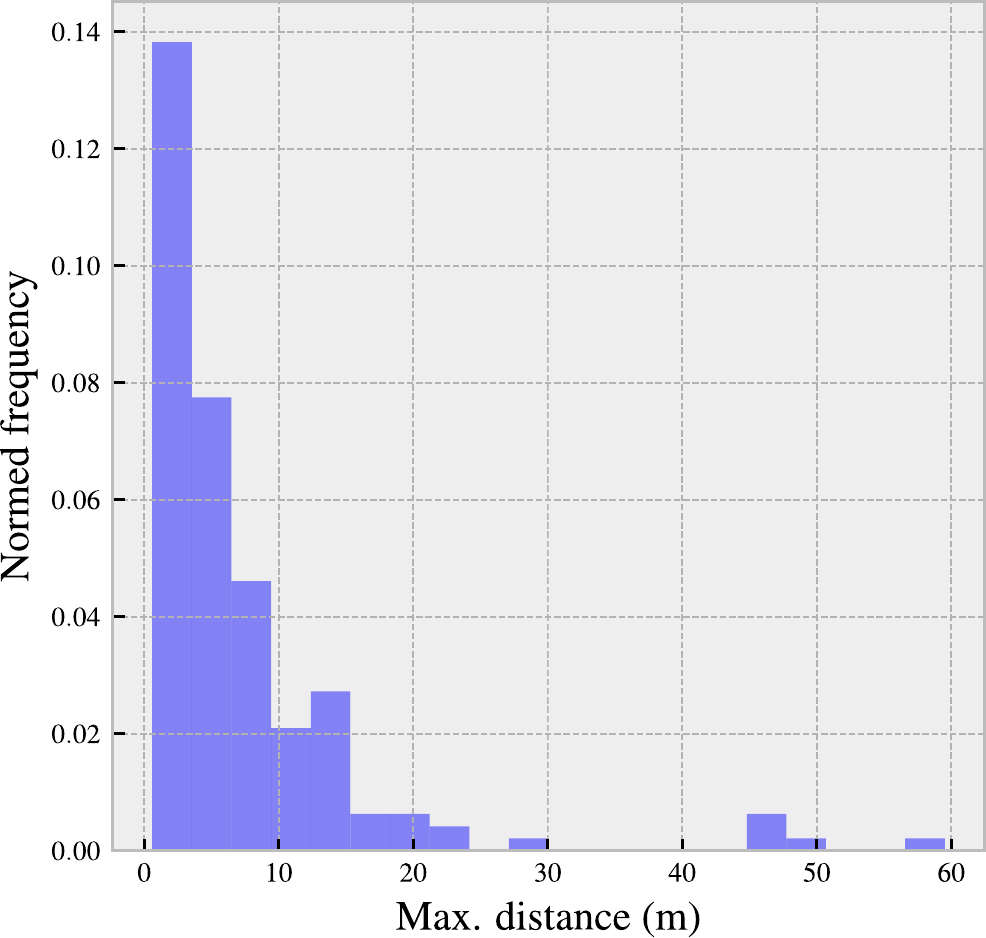}
	\caption[The maximum distance between the locations consisting of the loop state for $ k=1 $]{The maximum distance between the locations consisting of the loop state for $ k=1 $}
	\label{fig:max_dist_loop}
\end{figure}

\figPref\ref{fig:cmp_tf} shows the statistics of the \acs{tf}, denoted by the percentage of locations terminated with different conditions. In case of $ k=1 $ about 83\% iterative search processes have terminated with the converging state. This is about 35 percent points higher than in case of $ k=3 $. We therefore set the number of the nearest neighbors for \acs{knn} to 1. In addition, we have analyzed the searched locations within the looping state cases are distributed on space. \figPref\ref{fig:max_dist_loop} shows the distribution of the maximum distance between points within the same loop. Most of the maximum distances are significantly less than 10~m, though, in some extreme cases we observed up to 60~meters. As shown in \figPref\ref{fig:cmp_ecdf_opt}, the schemes proposed for the loop state (\acs{mcd} or \acs{mji}) are still capable of properly selecting position estimates close to the ground truth also in most of these cases.

\begin{figure}[!htb]
	\centering
	\subfloat[\acs{knn}]{\includegraphics[width=0.47\columnwidth]{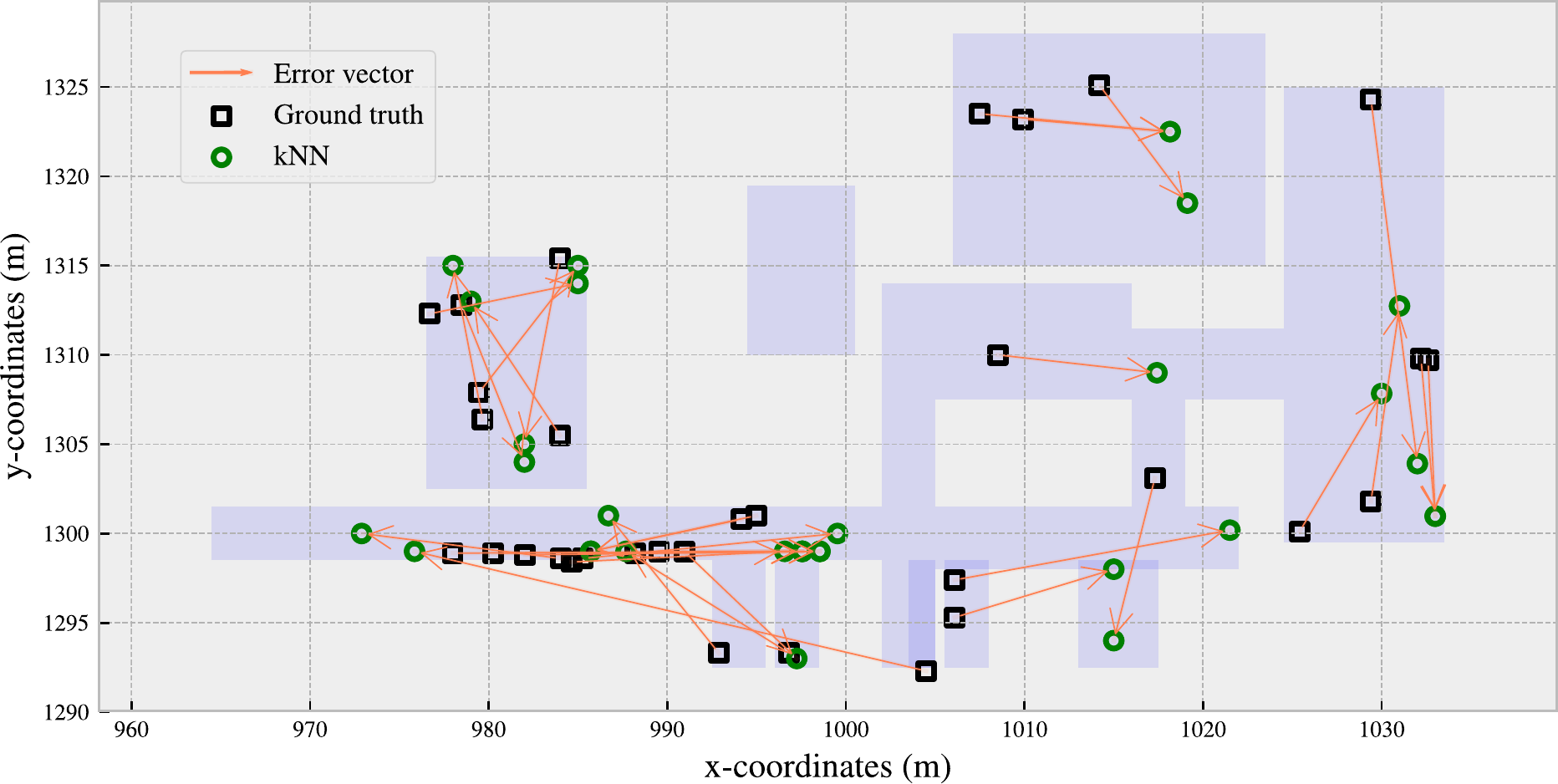}
		\label{subfig:distr_le_knn}}\hspace{2ex}
	\subfloat[Ours]{\includegraphics[width=0.47\columnwidth]{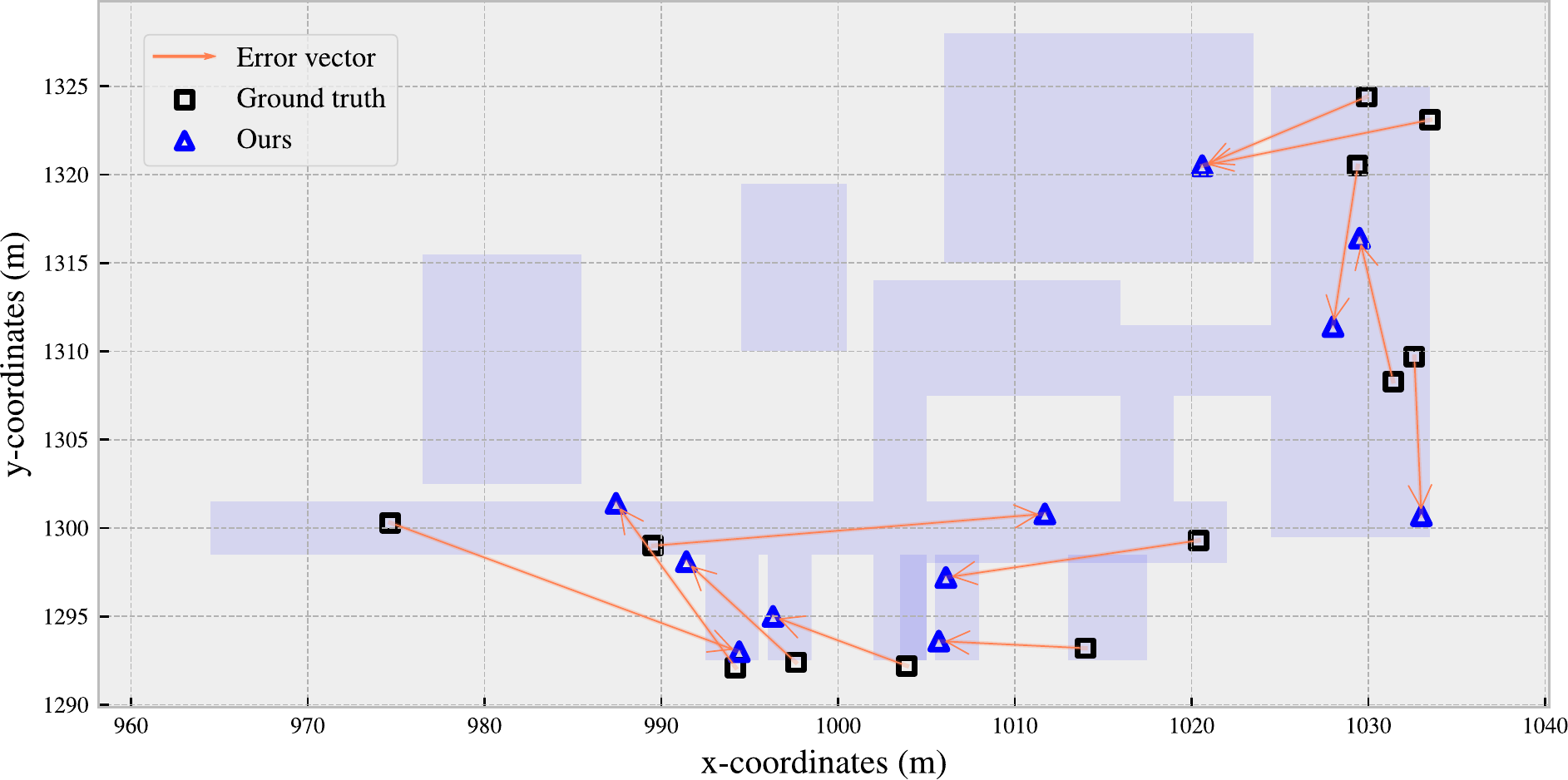}
		\label{subfig:distr_le_iter_knn}}
	\caption[Distribution of the locations yielding large errors ($ >8 $~m) in positioning]{Distribution of the locations yielding large errors ($ >8 $~m) in positioning}
	\label{fig:distr_le}
\end{figure}

The \acf{ecdf} of the radial positioning errors is presented in \figPref\ref{fig:cmp_ecdf_opt}. The proposed approach can significantly improve the positioning performance as compared to the performance of the traditional \acs{knn} and \acs{knn} with \acs{cdm}. Using the algorithm proposed herein, about 86\% of the estimated locations have a positioning error smaller than 2~m and around 97\% of estimated locations have an error of less than 4 m. Compared to \acs{knn}, this represent an improvement of up to 20 and 10 percent points, respectively. The improvement is also up to 10  and 6 percent points as compared to \acs{knn} with \acs{cdm}. The percentage of estimated locations whose error distance is larger than 5~m is reduced from about 10.2\% and 6.7\% to 2.6\%, when compared to the traditional \acs{knn}, and \acs{knn} with \acs{cdm}, respectively. We also report the \acf{ce} defined as the minimum radius for including a given percentage of positioning errors (\eg \acs{ce}~50 for the $ 50^{\texSym{th}} $ percentile) in \tabPref\ref{tab:stats_pe}. The maximum positioning error is reduced by about 40\%, from 37.0~m to 22.2~m when comparing \acs{knn} with \acs{cdm} to our approach. The \acs{ce}~50, \acs{ce}~75, and \acs{ce}~90 are reduced by one third when compared to the \acs{knn} without iterative positioning. Furthermore, in \figPref\ref{fig:distr_le} we illustrate and compare the distribution of the locations, at which the positioning error is larger than 8~m using the original \acs{knn}. \figPref\ref{subfig:distr_le_knn} shows that these locations yielding large errors are mostly located close to the accessible boundaries of the indoor regions, \ie close to corners of corridors and rooms, or to the walls. This pattern is similar to the spatial distribution of high variance of the feature values contained in the raw \acs{rfm} as shown in \figPref\ref{fig:rob_std}. Our approach can significantly reduce the number of occurrences where the positioning errors are larger than 8~m.

\begin{table}[!htb]
	\centering
	\caption[Radial positioning errors (meters)]{Statistics of positioning errors (meters)}
	\label{tab:stats_pe}
	\begin{tabular}{lllll}
		\hline\smallskip
		& \acs{ce}~50& \acs{ce}~75& \acs{ce}~90&Max. error\\
		\hline
		\acs{knn} &1.4 & 2.6& 5.2& 29.6\\
		\acs{knn} (\acs{cdm}) &1.1 & 1.9& 3.7& 37.0\\
		Ours &1.0&1.5&2.2&22.2\\
		Ours (opt.) &0.9&1.4&1.9& \,\,\,9.2\\
		\hline
	\end{tabular}
\end{table}

\section{Conclusion}\label{sec:conclusion}
We have proposed an iterative scheme for \acl{fbp}, which is based on the weighted dissimilarity measure, for reducing large errors occurring in \acsp{fips}. Appropriate weights for the individual feature can be obtained by analyzing the variability of the  kinematically collected raw data underlying the \acs{rfm}. The location-wise standard deviation of each feature is robustly computed using the \acs{mad} between the raw data and the spatially smoothed  \acs{rfm}. This variability information is stored as an additional layer of the \acs{rfm} and used for weighting the contribution of each feature to the dissimilarity measure during the online positioning phase.

Using real WLAN RSS data collected along with location ground truth in an office building, we could show that the noise of the raw observations indeed depends on the location and on the feature. We have implemented the proposed algorithms in Python and have validated the performance of the proposed iterative scheme. Compared to \acs{knn} with \acs{cdm}, the maximum positioning error is reduced by more than 40\% and the iterative scheme can improve the overall positioning performance. The positioning accuracy defined as the percentage of the locations whose radial positioning error is less than 2~m is improved from 65\% to 86\% when compared to traditional \acs{knn}. 

In future work, we will further investigate the proposed algorithms using data from other environments. We will further investigate the loop state and the handling of remaining outliers. Finally, we will investigate how the standard deviations modeled within the \acs{rfm} can help to identify the need for updates of the \acs{rfm}.

\section*{Acknowledgment}
The Chinese Scholarship Council has supported C.~Zhou during his doctoral studies at ETH Z\"urich. The data used within the experimental investigation were collected by the students E.~Weiss, I.~Bai, N.~Meyer, and G.~Filella. The device holder for the ground truth measurements was designed by R.~Presl.
\bibliographystyle{apalike}
\bibliography{../reference/sensors_var_iter_pos}



\end{document}